\documentclass{article}

\PassOptionsToPackage{numbers,sort&compress}{natbib}
\usepackage[main, final]{neurips_2026}
\usepackage[utf8]{inputenc} 
\usepackage[T1]{fontenc}    
\usepackage{hyperref}       
\usepackage{url}            
\usepackage{booktabs}       
\usepackage{amsfonts}       
\usepackage{nicefrac}       
\usepackage{microtype}      
\usepackage{xcolor}         
\usepackage{mathtools}
\usepackage{tabularx}
\usepackage{amsmath}
\usepackage{amssymb}

\usepackage{multirow}
\usepackage{subfig}
\usepackage{enumitem}
\usepackage{colortbl}
\usepackage{wrapfig}
\usepackage{todonotes}
\usepackage[most]{tcolorbox}
\newtcolorbox{takeaway}{
  colback=blue!3,     
  colframe=blue!70!black, 
  arc=2mm,            
  boxrule=0.8pt       
}
\usepackage{arydshln}
\usepackage{algorithm}
\usepackage{algpseudocode}
\usepackage{ulem}
\definecolor{yellow}{RGB}{255,192,0}
\definecolor{darkred}{RGB}{192, 0, 0}
\definecolor{beige}{RGB}{250,240,210}
\definecolor{peach}{RGB}{255,220,180}
\definecolor{lightpink}{RGB}{255, 235, 240}
\definecolor{softblue}{RGB}{0, 160, 220}
\definecolor{deepgreen}{RGB}{0,128,0} 
\definecolor{deepgray}{gray}{0.6}
\definecolor{midgray}{gray}{0.8}
\definecolor{lightgray}{gray}{0.87}
\usepackage[compatibility=false]{caption}
\usepackage{caption} 
\usepackage{subcaption}
\newcommand{\rand}
{\textit{Random}}
\newcommand{\cosine}
{\textit{Query}}
\newcommand{\ech}{\textit{\textbf{EchoPrune}}}
\newcommand{\mi}{MI-Pruner}

\newcommand{\say}[1]{\textcolor{gray}{\textit{#1}}}
\newcommand{\vv}[1]{\textcolor{gray}{\texttt{#1}}}

\newcommand{\tbd}[1]{\textcolor{red}{\textbf{#1}}}

\newcommand{\sota}[1]{{\textbf{#1}}}

\newcolumntype{Y}{>{\centering\arraybackslash}X}
\usepackage[table]{xcolor}

\title{\ech: Interpreting Redundancy as Temporal Echoes for Efficient VideoLLMs}

\author{
Jiameng Li$^1$~~~~
Minye Wu$^1$~~~~Jiezhang Cao$^2$~~~~Aleksei Tiulpin$^3$~~~~Matthew B. Blaschko$^1$\\
$^1$KU Leuven, ~~$^2$Shanghai Jiao Tong University,
~~$^3$Weill Cornell Medicine
\\
\textcolor{darkred}{
\url{https://github.com/renaissanceee/EchoPrune}}
}

\begin{document}

\maketitle

\begin{figure}[h]
\centering
\vspace{-0.5cm}
\includegraphics[width=\linewidth]
{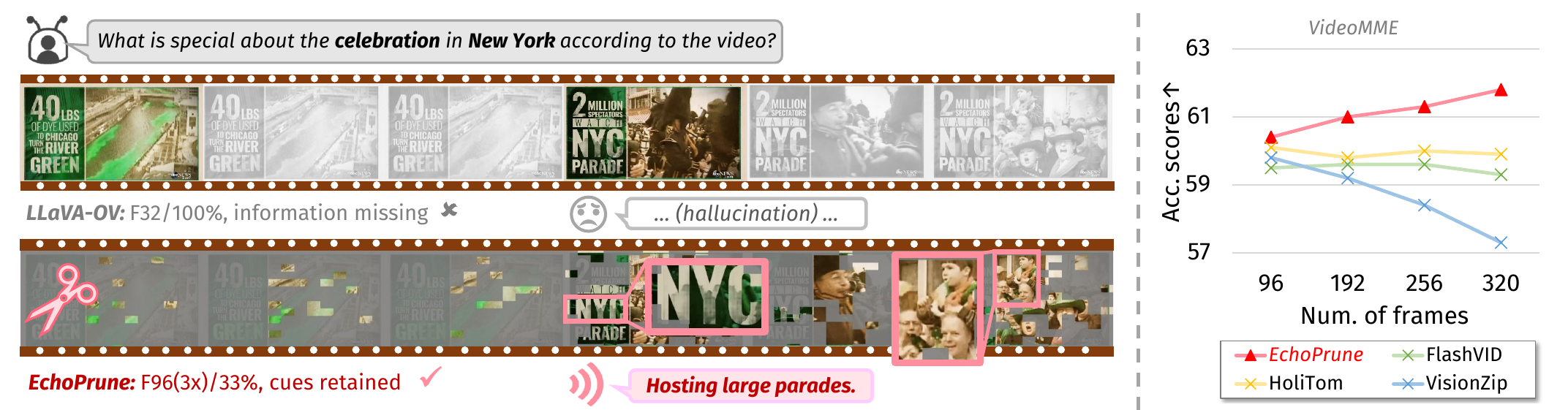} 
\caption{
\textbf{Left:} 
EchoPrune extends the temporal resolution with more visible frames for fine-grained video understanding, then selects pivot tokens to fit the budgets (F32 is short for 32 frames).
\textbf{Right:} Under the default token footprint, EchoPrune scales visible frames up to boost SOTA performance. 
}
\label{fig:fig1}
\end{figure}

\begin{abstract}

Long-form video understanding remains challenging for Video Large Language Models (VideoLLMs), as the dense frame sampling introduces massive visual tokens while sparse sampling risks missing critical temporal evidence and leading to LLM hallucination. Existing training-free token reduction methods either treat videos equally as static images or rely on segment-level merging heuristics, which weaken fine-grained spatiotemporal modeling and introduce additional overhead. In this paper, we propose \ech, a lightweight and training-free token pruning method that improves temporal resolution under a fixed LLM-side visual token budget. Our core idea is to interpret redundant video tokens as temporal echoes: if a token is well reconstructed from the previous frame, it is merely a temporally redundant echo; otherwise, it may capture new events, motion, or query-relevant visual evidence. Based on this insight, EchoPrune scores visual tokens by \textit{(i)} query-guided crossmodal relevance and \textit{(ii)} temporal reconstruction error, measured by correspondence matching and echo matching across consecutive frames. The selected tokens preserve task-relevant cues and temporal novelty while suppressing predictable redundancy, allowing VideoLLMs to observe more frames without increasing the decoding budget. Extensive experiments on LLaVA-OV, Qwen2.5VL, and Qwen3VL across six video understanding benchmarks show that EchoPrune enables VideoLLMs to process up to $\mathbf{20\times}$ frames under the same token budget, yielding improved performance ($\mathbf{5\%^\uparrow}$) and inference speedup ($\mathbf{5.6\times}$ for prefilling) on LLaVA-OV-7B.

\end{abstract}

\section{Introduction}
\label{sec:intro}

Video understanding has increasingly benefited from the integration of large language models, equipping Video Large Language Models (VideoLLMs) with strong multimodal reasoning capabilities \citep{qwen2.5-VL,li2024llavaov}. In VideoLLMs, a vision encoder \citep{zhai2023sigmoid} transforms sampled video frames into token sequences for LLM decoding.
However, such a dense tokenization introduces substantial redundancy, as consecutive frames exhibit high temporal correlation.
The resulting token proliferation significantly increases computational costs and limits practical deployments \citep{ji2025specvlm}.
Therefore, eliminating redundant video tokens while preserving task-relevant semantic information is critical for improving both the efficiency and applicability of VideoLLMs \citep{shao2025holitom,fan2026flashvid}.

Existing visual token reduction methods~\citep{jin2026compression,yanvision,shen2025fastvid} can be categorized into generic branches that treat images and videos equally, and video-specific ones considering the inductive bias in video data. 
Generic methods select tokens based on attention salience \citep{yang2025visionzip} or subset coverage \citep{li2026mi}, while video-specific methods exploit temporal structures via segment-level pruning \citep{shen2025fastvid}, token merging \citep{shao2025holitom}, uniqueness estimation~\citep{yuan2026unicomp}, or spatiotemporal organization~\citep{fan2026flashvid}. 
Despite their progress, these methods often rely on indirect or fixed criteria, such as miscalibrated attention~\citep{xiao2023efficient,yu2024unveiling}, predefined partitions or global pooling~\citep{shen2025fastvid,shao2025holitom,liu2025vidcom2}, and non-adaptive uniqueness measures~\citep{yuan2026unicomp}. 
Such designs can disrupt temporal continuity, weaken inter-segment information exchange, and discard fine-grained temporal cues that are essential for video understanding. 
As a result, they may suffer from temporal information loss and performance degradation, especially on long videos where relevant evidence is sparsely distributed over time \citep{li2026keeping}. 
Moreover, some of these methods \citep{shen2025fastvid} adopt DPC-KNN  clustering \citep{rodriguez2014clustering} or two-stage pipelines \citep{fan2026flashvid} from projection-layer pruning followed by inner-LLM merging, which introduces non-negligible computational overhead with implementation difficulty.

The central challenge is to mitigate the information bottleneck when representing abundant but unevenly distributed video content with compact visual tokens \citep{shao2025holitom,fan2026flashvid,yuan2026unicomp}. 
Sparse frame sampling is commonly adopted to avoid prohibitive token growth \citep{li2024llavaov}. However, it may discard critical visual evidence and introduce a semantic gap, causing a blind LLM to rely on language priors and hallucinate. This is illustrated in Fig.~\ref{fig:fig1} (left), where F32 means uniformly sampling 32 frames from a video clip along the temporal axis. This suggests that video pruning should not be treated merely as a means of reducing computation. Instead, the designed pruning is supposed to improve the temporal resolution \citep{fan2026flashvid,ju2026forestprune}, allowing the model to observe more frames under the same LLM-side token budget. 
Meanwhile, token importance in multimodal video understanding is inherently query-dependent \citep{song2025moviechat+,du2026unified}.
Therefore, an effective pruning strategy ought to preserve query-relevant evidence, capture newly emerging visual information, and suppress repetitive or predictable content with minimal compression overhead.

In this paper, we propose \ech, a query-aware and lightweight video token pruning method that preserves informative visual evidence while suppressing temporal redundancy. 
Our key insight is that redundant video tokens can be interpreted as \textit{temporal echoes}: if a token in the current frame can be well predicted by tokens from the preceding frame, it is likely to carry repetitive information; otherwise, it may correspond to newly emerging events, local motion, or task-relevant visual evidence. 
To operationalize this idea, we formulate video token pruning under the framework of \textit{Maximum Marginal Relevance (MMR)} \citep{carbonell1998use}, where the importance of each token is determined by both its relevance to the user query and its redundancy regarding preceding visual context.
Specifically, EchoPrune scores each video token from three complementary perspectives. 
First, \textit{query-guided crossmodal relevance} measures the semantic alignment between visual tokens and the textual query, encouraging the model to preserve visual evidence needed for answering the question. 
Second, \textit{correspondence matching} captures local content changes by comparing tokens across adjacent frames, which helps identify motion-related visual information. 
Third, \textit{echo matching} reconstructs the current token from a spatial neighborhood in the preceding frame, providing a lightweight proxy for temporal predictability. These two types of matching form the redundancy proxy called \textit{temporal reconstruction error}.
By integrating these scores, EchoPrune performs once-for-all Top-K pruning before the LLM, without additional training or intrusive modifications to the pretrained VideoLLM. As shown in Fig.~\ref{fig:fig1} (right), under the same token budget, EchoPrune continues benefiting from using more frames, whereas competing methods tend to saturate or degrade, indicating richer temporal evidence preservation for stronger long video understanding. 

Our key contributions are summarized as follows:
\begin{itemize}[nosep,leftmargin=0.6cm]
    \item We propose \ech, a training-free and lightweight video token pruning method that preserves task-relevant visual evidence while suppressing temporal redundancy.

    \item We interpret video redundancy as \textit{temporal echoes} and combine query-guided relevance, correspondence matching, and echo matching for a holistic Top-K token selection (Sec.~\ref{sec:methods}).

    \item Extensive experiments on LLaVA-OV, Qwen2.5VL, and Qwen3VL show that \ech~supports up to $20\times$ more frames under the same token budget, achieves SOTA performance, and accelerates inference with up to $5.6\times$ prefilling and $2.1\times$ TTFT speedup on LLaVA-OV (Sec.~\ref{sec:exp}).
\end{itemize}

\section{Preliminaries}
\label{sec:pre}

\paragraph{Pruning stages.}
Grounded in the theory of information bottleneck \citep{tishby2000information}, many pruning methods intervene in the projection layer \citep{shen2025fastvid,yuan2026unicomp}, where multimodal tokens are aligned in a shared embedding space but have not yet fused through cross-attention. Meanwhile, empirical evidence from FrameFusion \citep{fu2025framefusion} suggests that the ranking of token similarity hardly changes across LLM layers. 
While HoliTom \citep{shao2025holitom} and FlashVID \citep{fan2026flashvid} introduce inner-LLM merging to further exploit cross-modal interactions in LLMs, this two-stage pipeline inevitably increases computational overhead and implementation complexity.
To achieve \textit{once-for-all} effects, our pruning is conducted once in projection space like FastVID and UniComp \citep{shen2025fastvid,yuan2026unicomp}.

\paragraph{Maximum marginal relevance (MMR).} 
Video token pruning can be viewed as a subset selection problem, aiming to find a representative subset maximizing information coverage. Specifically, the pruning objective can be characterized as a submodular function which satisfies the \textit{diminishing returns} property \citep{fujishige2005submodular}, \textit{i.e.}, the incremental benefit of incorporating a new video token decreases as the selected set grows. This mathematical foundation motivates our adoption of the MMR framework \citep{carbonell1998use}, whose scoring function incorporates both \underline{r}elevance terms for task-oriented importance and re\underline{d}undancy terms to capture the diminishing gain. 
Among video pruning methods, MMG-Vid \citep{ma2026mmg} calculates coarse frame-level MMR for frame budgets. SemVID \citep{li2026keeping} derives MMR scores as the query evidence and intra-frame redundancy, while its greedy search relies on predefined segments and an anchor set.
By contrast, our EchoPrune formulates the redundancy term as the reconstruction error from the history frame and conducts a single holistic ranking, which yields better efficiency.

\paragraph{Frame matching and reconstruction.}
As a cornerstone of computer vision, frame matching has been adapted in token pruning for VideoLLMs. MMG-Vid \citep{ma2026mmg} studies frame-level similarity, and subsequent methods shift towards token-level uniqueness to analyze temporal dependencies across consecutive frames.
Denote $\mathbf{v}_i^k$ as the $i_\text{th}$ token in frame $k$, 
SemVID \citep{li2026keeping} and KiToke \citep{huang2026kitoke} generally quantify temporal connectivity via the $L_2$ distance, written as $L_2(\mathbf{v}_i^{k},\mathbf{v}_i^{k-1})$.
KiToke \citep{huang2026kitoke} further evaluates spatial displacement by the minimum $L_2$ distance between a token and all candidates in the preceding frame, \textit{i.e.}, $\min L_2(\mathbf{v}_i^{k},\mathbf{v}_j^{k-1})$. 
In comparison, we replace the minimal aggregation with echo matching to preserve fine-grained semantics.

\section{Methods}
\label{sec:methods}

\begin{figure}[t]
\centering
\includegraphics[width=\linewidth]{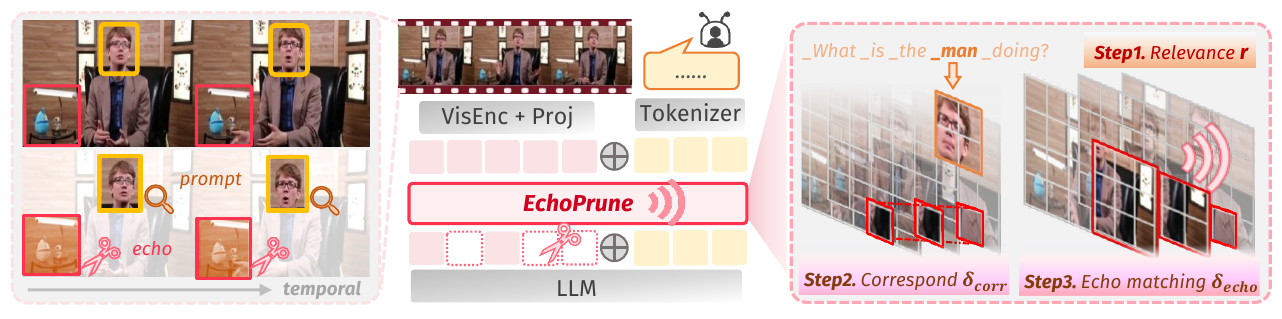} 
\vspace{-0.4cm}
\caption{
\textbf{Overview.} 
EchoPrune identifies the most informative tokens via a three-fold decomposition:
\textbf{Step 1.} Crossmodal relevance guided by user query ($r$);
\textbf{Step 2.} Spatial motion based on correspondence matching ($\delta_\mathrm{corr}$);
\textbf{Step 3.} Temporal novelty driven by echo matching ($\delta_\mathrm{echo}$).
}
\vspace{-0.2cm}
\label{fig:pipe}
\end{figure}

Video tokens are selected according to their MMR scores following the procedure in Algorithm \ref{alg:pipe}. 
As shown in Fig.~\ref{fig:pipe} (left), the narrator{\setlength{\fboxrule}{1pt}\setlength{\fboxsep}{2.5pt}\fcolorbox{yellow}{beige}{}} in the video is essential to answer the question \texttt{"What is the man doing?"} (relevance), while the background lamp{\setlength{\fboxrule}{1pt}\setlength{\fboxsep}{2.5pt}\fcolorbox{red}{lightpink}{}} persists across frames with less temporal information (redundancy). This observation motivates our MMR-based token selection.
In the task of Visual Question Answering (VQA), a straightforward MMR score of token $\mathbf{v}_i$ can be written as:
\begin{align}
S_i &= R(\mathbf{v}_i,\mathcal{T})-D(\mathbf{v}_i,\mathcal{V}_\mathrm{S}).
\label{eq:mmr}
\end{align}
Here, the relevance term ($R$) aggregates over the text set $\mathcal{T}$, while the redundancy term ($D$) measures the marginal contribution to the selected vision set $\mathcal{V}_\mathrm{S}$.
To estimate the \textbf{redundancy}, EchoPrune models the current frame as a \textit{temporal echo} of its predecessor. By causally propagating the information from earlier to later frames, we derive the reconstruction error between adjacent frames as a self-supervised metric to filter out predictable (\textit{i.e.}, replaceable) tokens. 
Our reconstruction is guided by two complementary concerns: \textit{(i) temporal novelty:} tokens in frame $k$ that are poorly reconstructed from the last frame tokens $\mathcal{V}^{k-1}$ indicate the arrival of a new event with semantic importance; \textit{(ii) spatial motion:} a substantial semantic shift for inter-frame tokens at the same position suggests the presence of dynamic objects. Consequently, we calculate reconstruction error through echo matching ($\delta_\mathrm{echo}$) and correspondence matching ($\delta_\mathrm{corr}$) to quantify the temporal novelty and spatial motion. The \textbf{relevance} term is defined as the maximum aggregation from crossmodal relevance ($r$). Our pipeline is illustrated in Fig.~\ref{fig:pipe}. Let $\mathbf{v}_i^k$ denote a token in frame $k$ at position $i$.
Following Eqn.~(\ref{eq:mmr}), EchoPrune frames the MMR score of video tokens $\{\mathbf{v}_i^k\}$ as:
\begin{align}
R(\mathbf{v}_i^k,\mathcal{T}) &\coloneqq  r(\mathbf{v}_i^k, \mathcal{T})
\\
D(\mathbf{v}_i^k,\mathcal{V}^{k-1}) &\coloneqq \delta_\mathrm{corr}(\mathbf{v}_i^k,\mathcal{V}^{k-1})
+
\delta_\mathrm{echo}(\mathbf{v}_i^k,\mathcal{V}^{k-1}).
\end{align}
The definitions of these reconstruction errors are presented as follows. 

\subsection{Crossmodal Relevance}
\label{subsec:method_rele}

The query-guided crossmodal relevance term is formulated as an attention-like interaction between a normalized visual token $\tilde{\mathbf{v}}_i^k$ and a normalized textual token $\tilde{\mathbf{t}}_j$. 
As observed in MI-Pruner \citep{li2026mi}, the average pooling tends to flatten the semantic nuance. To preserve the most salient crossmodal signals, we employ maximum aggregation among text tokens $\tilde{\mathcal{T}}$:
\begin{align}
r\left(\tilde{\mathbf{v}}^{k}_i,\tilde{\mathcal{T}}\right) &= \max_{\tilde{\mathbf{t}}_j\in \tilde{\mathcal{T}}} (\tilde{\mathbf{v}}_i^{k})^{\top} \tilde{\mathbf{t}}_j,
\label{eq:rel_cross}
\end{align}
where $\tilde{\cdot}, {\cdot}^\top$ denote $L_2$ normalization and transposition.
The relevance term measures the token's multimodal significance in the VQA task. We show its importance in the ablation study (Sec.~\ref{subsec:exp_ablation}).

\subsection{Temporal Reconstruction}
\label{subsec:method_recon}

We evaluate the video redundancy through two types of reconstruction error, $\delta_{\mathrm{corr}}$ and $\delta_{\mathrm{echo}}$. 
To establish a temporal correspondence of $\mathbf{v}_i^k$, we first define the \textit{correspondence reconstruction error} $\delta_{\mathrm{corr}}$ to directly measure the temporal persistence at the same position $i$:
\begin{align}
\delta_{\mathrm{corr}}\left(\tilde{\mathbf{v}}^{k}_i, \tilde{\mathcal{V}}^{k-1}\right) &= (\tilde{\mathbf{v}}_i^k)^\top \tilde{\mathbf{v}}_i^{k-1}.
\label{eq:err_id}
\end{align}
The negative sign is deferred to the final score Eqn.~(\ref{eq:score}).
Then, we construct a similarity matrix between $\mathbf{v}_i^k$ and last-frame candidates $\{\mathbf{v}_j^{k-1}\}$. Following the energy-based formulation, the temperature scaled similarity score $\boldsymbol{\rho}^{k}_{ij}$ is written as:
\begin{align}
    \boldsymbol{\rho}^{k}_{ij} &= \frac{(\tilde{\mathbf{v}}_i^{k})^{\top} \tilde{\mathbf{v}}^{k-1}_j}{\tau},
\end{align}
\begin{wrapfigure}{r}{0.5\textwidth} 
  \centering
  \vspace{-0.2cm}
  \includegraphics[width=\linewidth]{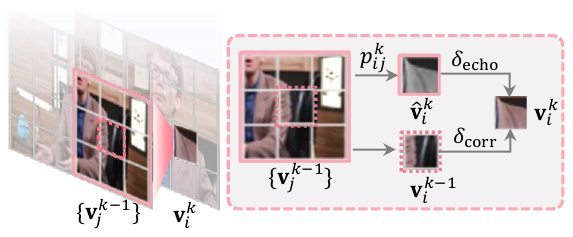} 
  \vspace{-0.4cm}
  \caption{Temporal reconstruction ($\tilde{\cdot}$ omitted).}
  \label{fig:recon}
\end{wrapfigure}
where $\tau$ serves as the temperature parameter. 
To enforce local spatiotemporal coherence and reduce computational overhead, we restrict the candidate set of $\{\mathbf{v}_j^{k-1}\}$ to a spatial neighborhood $\Omega_i^\epsilon$ centered at position $i$ with window size $\epsilon$, \textit{e.g.}, $3{\times}3$.
Then we apply softmax over candidates from frame $k{-}1$ to obtain the conditional probability $p_{ij}^k$:
\begin{align}
    p_{ij}^k &\triangleq p\left(\tilde{\mathbf{v}}^{k-1}_j | \tilde{\mathbf{v}}_i^{k}\right) = \mathop{\mathrm{softmax}}\limits_{j \in \Omega_i^\epsilon} \left(\boldsymbol{\rho}^{k}_{ij}\right),
    \label{eq:prob}
\end{align}
which serves as the matching weight for the temporal echo.
The token $\tilde{\mathbf{v}}_i^k$ can be approximated (denoted as $\hat{\tilde{\mathbf{v}}}_i^k$) by a linear combination of its temporal neighbors in frame $k{-}1$:
\begin{align}
\hat{\tilde{\mathbf{v}}}_i^k &= \sum_{j \in \Omega_i^\epsilon} p_{ij}^k \tilde{\mathbf{v}}_j^{k-1}.
\label{eq:recon}
\end{align}
Consistent with Eqn.~(\ref{eq:err_id}), we define the \textit{echo reconstruction error} $\delta_{\mathrm{echo}}$ as the inner product between the original token and its reconstructed counterpart:
\begin{align}
\delta_{\mathrm{echo}}\left(\tilde{\mathbf{v}}^{k}_i, \tilde{\mathcal{V}}^{k-1}\right) = (\tilde{\mathbf{v}}_i^k)^\top \hat{\tilde{\mathbf{v}}}_i^k = (\tilde{\mathbf{v}}_i^k)^\top \bigg(\sum_{j \in \Omega_i^\epsilon} p_{ij}^k \tilde{\mathbf{v}}^{k-1}_j \bigg).
\label{eq:err_se}
\end{align}
As shown in Fig.~\ref{fig:recon}, the overall reconstruction error combines these two components:
\begin{align}
\delta &= \delta_{\mathrm{corr}}+\delta_{\mathrm{echo}}.
\label{eq:err}
\end{align}

\begin{algorithm}[h]
\caption{EchoPrune}
\label{alg:pipe}
\begin{algorithmic}[1]

\State \textbf{Input:} 
\small{
Dense video tokens $\mathcal{V}{=}\{\mathbf{v}_{i}^{k}\}$, $|\mathcal{V}|{=}\gamma N_V$; text tokens $\mathcal{T}{=}\{\mathbf{t}_{j}\}_{j=1}^{N_T}$; temperature $\tau$; window size $\epsilon$.
}

\State \textbf{Output:}
Selected video tokens $\mathcal{V}_\mathrm{S}~\text{, where}~{|\mathcal{V}_\mathrm{S}|{=}N_V}$.

\State\textcolor{gray}{\textit{// Step 1. Crossmodal relevance}}
\State $r(\tilde{\mathbf{v}}^{k}_i,\tilde{\mathcal{T}})\leftarrow \max_{\tilde{\mathbf{t}}_j\in \tilde{\mathcal{T}}} (\tilde{\mathbf{v}}_i^{k})^\top \tilde{\mathbf{t}}_j$ 

\State\textcolor{gray}{\textit{// Step 2-3. Intra-modal reconstruction error (redundancy)}}

\State $\boldsymbol{\rho}^{k}_{ij} \leftarrow \tilde{\mathbf{v}}_i^{k\top} \tilde{\mathbf{v}}^{k-1}_j / \tau$
\Comment{Scaled similarity}

\State {$p_{ij}^k \triangleq p(\tilde{\mathbf{v}}^{k-1}_j | \tilde{\mathbf{v}}_i^{k}) \leftarrow \text{softmax}_{j \in \Omega_i^\epsilon}(\boldsymbol{\rho}^{k}_{ij})$}
\Comment{Conditional probabilities}

\State {$\delta(\tilde{\mathbf{v}}^{k}_i,\tilde{\mathcal{V}}^{k-1}) \leftarrow 
(\tilde{\mathbf{v}}_i^k)^\top\tilde{\mathbf{v}}^{k-1}_i+(\tilde{\mathbf{v}}_i^k)^\top(\sum_{j \in \Omega_i^\epsilon} p_{ij}^k \tilde{\mathbf{v}}^{k-1}_j) $ } 
\Comment{Reconstruction error}

\State\textcolor{gray}{\textit{// MMR selection}}
\State $S(\tilde{\mathbf{v}}^{k}_i,\tilde{\mathcal{V}},\tilde{\mathcal{T}}) \leftarrow r(\tilde{\mathbf{v}}^{k}_i,\tilde{\mathcal{T}})- \delta(\tilde{\mathbf{v}}^{k}_i,\tilde{\mathcal{V}}^{k-1})$

\State $\mathcal{V}_\mathrm{S}{=}\{\mathbf{v}_i^k\}, ~\text{with}~{|\mathcal{V}_\mathrm{S}|{=}N_V} \leftarrow \mathop{\operatorname{Top-K}_{N_V}}  (S(\tilde{\mathbf{v}}^{k}_i,\tilde{\mathcal{V}},\tilde{\mathcal{T}}))$ 

\State \Return $\mathcal{V}_\mathrm{S}$.

\end{algorithmic}
\end{algorithm}

\subsection{Overall MMR Scores}
\label{subsec:method_all}

Given Eqn.~(\ref{eq:rel_cross}) and Eqn.~(\ref{eq:err}), the MMR score of EchoPrune is written as:
\begin{align}
S\left(\tilde{\mathbf{v}}^{k}_i,\tilde{\mathcal{V}},\tilde{\mathcal{T}}\right) &= r\left(\tilde{\mathbf{v}}^{k}_i,\tilde{\mathcal{T}}\right)- \delta\left(\tilde{\mathbf{v}}^{k}_i,\tilde{\mathcal{V}}^{k-1}\right).
\label{eq:score}
\end{align}
Under the token budget $B$, a simple Top-K sorting is sufficient to obtain the selected set $\mathcal{V}_\mathrm{S}$.
\begin{align}
\mathcal{V}_\mathrm{S} &= \mathop{\operatorname{Top-K}_{B}}  \left(S(\tilde{\mathbf{v}}^{k}_i,\tilde{\mathcal{V}},\tilde{\mathcal{T}})\right).
\label{eq:top}
\end{align}

We denote $N_V{=}M_F{\cdot} N_F$ as the default budgets, where $N_V$ consists of $M_F$ frames with $N_F$ tokens/frame, \textit{e.g.}, LLaVA-OV \citep{li2024llavaov} sets $M_F{=}32$ frames and $N_F{=}196$ tokens/frame. To improve the temporal resolution under the default budget $N_V$, we retain $\frac{1}{\gamma}$ tokens from the extended $\gamma M_F ~(\gamma>1)$ frames, as shown in Algorithm~\ref{alg:pipe}.  
Since the first frame has no reconstruction error, we retain an average of $\frac{N_F}{\gamma}$ tokens based on its crossmodal relevance.

\section{Experiments}
\label{sec:exp}

\subsection{Setup}
\label{subsec:exp_setup}

\paragraph{Benchmarks.}
We evaluate our method on 6 video understanding benchmarks: VideoMME \citep{fu2025video} (without subtitle), EgoSchema \citep{mangalam2023egoschema}, LongVideoBench \citep{wu2024longvideobench}, MLVU \citep{zhou2025mlvu}, VideoMMMU \citep{hu2025video} and LVBench \citep{wang2025lvbench}. The evaluation follows the standard settings of LMMs-Eval \citep{zhang2025lmms}. 
We adopt LLaVA-OV \citep{li2024llavaov}, instruction-tuned Qwen2.5VL \citep{qwen2.5-VL} and Qwen3VL \citep{Qwen3VL} as the pretrained VideoLLMs for token pruning. Results on various model scales and settings can be found in App. \ref{subsec:app_more_exp}. 
To ensure sufficient evidence frames,
we increase the temporal resolution with higher frame sampling rates. Specifically, we
scale the number of frames up to F320, while keeping the default LLM token budget fixed, \textit{i.e.}, selecting the same amount of tokens from extended frames. 
This setting avoids undersampling bias and LLM hallucination in F32 benchmarks \citep{shen2025fastvid,shao2025holitom}, where evidence frames may already be missed before pruning. 
See more discussions about temporal sampling in App.~\ref{subsec:app_related_sample}.

\paragraph{Comparison methods.} We reproduce six comparison methods using their official repositories, which are categorized into: 
\textit{(i) generic methods:} applicable for both image and video pruning, like attention salience guided VisionZip \citep{yang2025visionzip} and subset coverage based MI-Pruner \citep{li2026mi}; 
\textit{(ii) video-specific methods:} considering spatiotemporal processing, \textit{e.g.}, FastVID \citep{shen2025fastvid}, HoliTom \citep{shao2025holitom}, UniComp \citep{yuan2026unicomp} and FlashVID \citep{fan2026flashvid}. Descriptions of the comparison methods can be found in Sec.~\ref{sec:related}.

\textbf{Implementation.} We test on a single A100 GPU with \texttt{batch\_size}=1 following previous work \citep{fan2026flashvid,yuan2026unicomp}.
For Qwen3VL, we preserve the inter-frame timestamp tokens outside of token budgets.
The model architecture affects hyperparameters due to different tokenization strategies in visual encoders. Based on experimental results in the ablation study (Sec.~\ref{subsec:exp_ablation}), we select the optimal parameters for each model family and keep this set of parameters fixed in the comparative experiments.

\begin{table}[th]
\centering
\caption{\textbf{Performance on LLaVA-OV-7B.} Keeping the F32 budget, F96/33.3\% denotes keeping 33.3\% tokens from 96 frames.
EchoPrune shows SOTA performance on long video benchmarks represented by MLVU and LVBench. Publication venues or preprint dates are indicated in \vv{grey}.}
\label{tab:results_ov_32}
\resizebox{0.99\textwidth}{!}{
\begin{tabular}{lccccccccc|cc}
\toprule
\multirow{2}{*}{\textbf{Method}} & \multicolumn{4}{c}{\textbf{VideoMME}} & \multicolumn{2}{c}{\textbf{EgoSchema}} & \textbf{LongVideo} & \multirow{2}{*}{\textbf{MLVU}} & \multirow{2}{*}{\textbf{LVBench}}  & \multicolumn{2}{c}{\textbf{Average}} \\ 
 & Short & Medium & Long & Overall & Subset & Total & \textbf{Bench} & & & Scores & \% \\
Duration & $<$ 3 min & 3-30 min & $>$ 30 min & 1-60 min & $<$ 3 min & $<$ 3 min & 1-60 min  &  3-120 min &  30-120 min &  \multicolumn{2}{c}{$<$ 120 min} \\
\midrule
LLaVA-OV-7B & 70.1 & 56.7 & 48.9 & 58.6  & 62.2 & 60.3 & 56.4 & 47.0 & 38.4 & 52.1 & 100 \\
\hline
\rowcolor{lightgray}
\multicolumn{12}{c}{\textit{F96 (3$\times$) / 33.3\%}}
\\
\rowcolor{white}
MI-Pruner \citep{li2026mi} \vv{2026.04} &
70.7 & 60.7 & 49.0 & 60.1 & 62.8 & 60.5 & 55.3 & 48.0 & 39.3 & 52.6 & +1.0 \\
VisionZip \citep{yang2025visionzip} \vv{CVPR'25} &
71.7 & 57.9 & 49.8 & 59.8  & 64.2 & 60.9 & 57.5 & 46.8 & 39.7 & 52.9 & +1.5 \\
FastVID \citep{shen2025fastvid} \vv{NeurIPS'25} &
 69.4 & 56.6 & 48.8 & 58.3  & 63.0 & 59.6 & 56.3 & 45.6 & 40.0 & 51.9 & -0.4 \\
HoliTom \citep{shao2025holitom} \vv{NeurIPS'25} &
\sota{73.0} & 58.2 & 49.1 & 60.1  & \sota{64.8} & 61.5 & 58.3 & 47.1 & 40.2 & 53.4 & +2.5 \\
UniComp \citep{yuan2026unicomp} \vv{CVPR'26} &
71.4 & \sota{59.4} & 49.6 & 60.1  & 63.6 & \sota{62.3} & \sota{58.5} & 47.5 & 40.8 & \sota{53.8} & \sota{+3.3} \\
FlashVID \citep{fan2026flashvid} \vv{ICLR'26} &
72.4 & 56.7 & 49.3 & 59.5  & 63.8 & 61.0 & 58.4 & 47.5 & 41.0 & 53.5 & +2.7 \\
\rowcolor{lightpink}
\ech
&
71.9 & 58.9 & \sota{50.5} & \sota{60.4}  & 62.3 & 60.7 & 58.2 & \sota{48.2} & \sota{41.3} & \sota{53.8} & \tbd{+3.3} \\

\hline
\rowcolor{lightgray}
\multicolumn{12}{c}{\textit{F256  (8$\times$) / 12.5\%}}\\
\rowcolor{white}
\mi~\citep{li2026mi} \vv{2026.04}
& 66.8 & 58.1 & 50.1 & 58.3  & 63.0 & 60.1 & 53.1 & 47.4 & 40.8 & 51.9 & -0.4 \\
VisionZip \citep{yang2025visionzip} \vv{CVPR'25}
& 68.4 & 56.8 & 50.1 & 58.4 & 63.5 & 61.7 & 54.7 & 47.0 & 40.4 & 52.4 & +0.6 \\
FastVID \citep{shen2025fastvid} \vv{NeurIPS'25}
& 64.6 & 54.4 & 48.8 & 55.9 & 59.4 & 57.3 & 52.3 & 42.7 & 39.8 & 49.6 & -3.1 \\
HoliTom \citep{shao2025holitom} \vv{NeurIPS'25}
& \sota{71.8} & 57.7 & 50.4 & 60.0 & 64.1 & 61.2 & 56.2 & 47.4 & 42.9 & 53.5 & +2.7 \\
UniComp \citep{yuan2026unicomp} \vv{CVPR'26}
& 70.7 & 60.6 & 50.2 & 60.5 & \sota{64.2} & \sota{62.4} & 57.4 & 49.5 & 43.1 & 54.6 & +4.8 \\
FlashVID \citep{fan2026flashvid}  \vv{ICLR'26}
& 71.1 & 57.1 & 50.7 & 59.6 & 63.4 & 60.1 & \sota{58.3} & 48.0 & 42.3 & 53.7 & +3.1 \\
\rowcolor{lightpink}
\ech
& \sota{71.8} & \sota{61.3} & \sota{50.8} & \sota{61.3} & 64.0 & 61.3 & 57.8 & \sota{49.7} & \sota{43.3} & \sota{54.7}  & \tbd{+5.0} \\
\hline
\rowcolor{lightgray}
\multicolumn{12}{c}{\textit{F320 (10$\times$) / 10\% \textbf{\say{(ultra)}}}} 
\\
\rowcolor{white}
\mi~\citep{li2026mi}  \vv{2026.04} & 65.8 & 57.4 & 49.1 & 57.4 & 61.8 & 59.9 & 51.0 & 45.9 & 40.0 & 50.8 & -2.5 \\
VisionZip \citep{yang2025visionzip}  \vv{CVPR'25} & 67.0 & 56.8 & 48.1 & 57.3  & 63.2 & 60.7 &  53.4 & 44.8 & 40.5 & 51.3 & -1.5  \\
FastVID \citep{shen2025fastvid} \vv{NeurIPS'25}  & 63.3 & 53.8 & 46.9 & 54.7 & 56.7 & 55.9 &  50.6 & 41.3 & 38.7 & 48.2 & -7.5 \\
HoliTom \citep{shao2025holitom}  \vv{NeurIPS'25} & \sota{72.7} & 56.8 & 50.2 & 59.9 & 63.8 & 61.1 &  56.5 & 48.3 & 42.0 & 53.6 & +2.9 \\
UniComp \citep{yuan2026unicomp}  \vv{CVPR'26} & 72.0 & 59.8 & 50.3  & 60.7  & \sota{64.4} & \sota{61.2} &  \sota{58.5} & 47.7 & 42.0 & 54.0 & +3.6 \\
FlashVID \citep{fan2026flashvid}  \vv{ICLR'26} & 70.7 & 56.7 & 50.4 & 59.3 & 63.4 & 60.8 &  57.0 & 47.6 & \sota{42.9} & 53.5 & +2.7 \\
\rowcolor{lightpink}
\ech
& 71.7 & \sota{61.1} & \sota{52.6} & \sota{61.8} & 63.0 & 60.4 & 57.6 & \sota{49.8} & \sota{42.9} & \sota{54.5}  & \tbd{+4.6} \\
\bottomrule
\end{tabular}
}
\end{table}

\begin{figure}[h]
\centering
\includegraphics[width=0.99\linewidth]{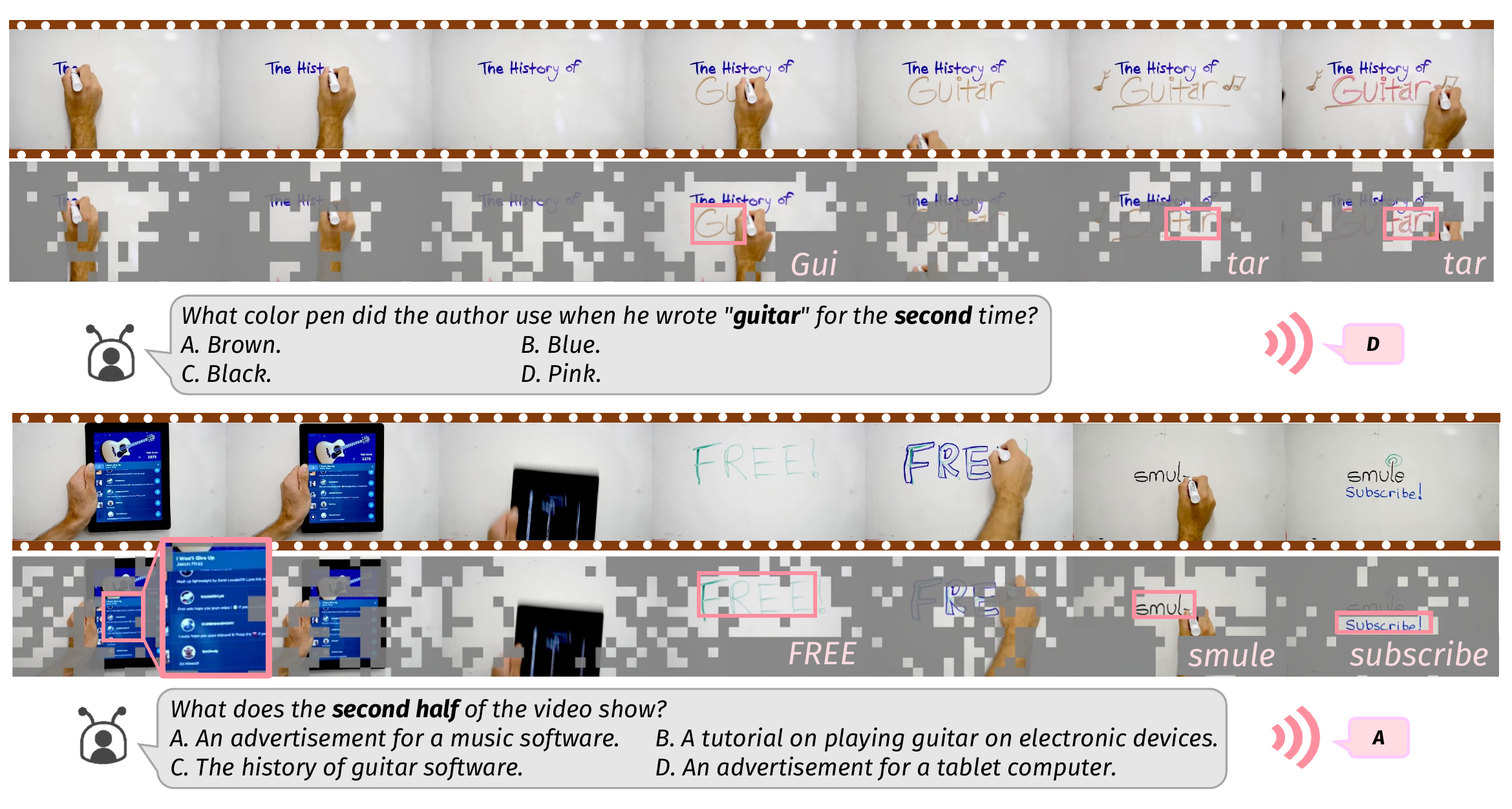} 
\caption{
\textbf{Pruning effects on Qwen2.5VL-7B}. EchoPrune preserves the count and OCR patch \texttt{"-tar"} and identifies the logo \texttt{"smule"} and contextual cues \texttt{"subscribe"}. Two questions are from the same video in VideoMME (F160/20\%). Pivot text cues are highlighted in \textbf{bold}.
}
\label{fig:vis_qwen2_5}
\end{figure}

\begin{table}[th]
\centering
\caption{\textbf{Performance on Qwen2.5VL-7B.} We keep F16 token budget to compare pruning effects following \cite{fan2026flashvid}. EchoPrune demonstrates consistent effectiveness in average scores.}
\label{tab:results_qwen2.5_16}
\resizebox{0.99\textwidth}{!}{
\begin{tabular}{lccccccccc|cc}
\toprule
\multirow{2}{*}{\textbf{Method}} & \multicolumn{4}{c}{\textbf{VideoMME}} & \multicolumn{2}{c}{\textbf{EgoSchema}} & \textbf{LongVideo} & \multirow{2}{*}{\textbf{MLVU}} & \textbf{Video} & \multicolumn{2}{c}{\textbf{Average}} \\ 
 & Short & Medium & Long & Overall & Subset & Total & \textbf{Bench} & & \textbf{MMMU} & Scores & \% \\
Duration & $<$ 3 min & 3-30 min & $>$ 30 min & 1-60 min & $<$ 3 min & $<$ 3 min & 1-60 min  &  3-120 min &  $<$ 20 min &  \multicolumn{2}{c}{$<$ 120 min} \\
\midrule
Qwen2.5VL-7B & 66.4 & 56.4 & 48.2 & 57.0 & 58.2 & 55.6 & 56.9 & 40.6 & 45.9 & 51.2 & 100 \\

\hline
\rowcolor{lightgray}
\multicolumn{12}{c}{\textit{F64 (4$\times$) / 25\%}} 
\\
\rowcolor{white}
\rand & 72.0 & 59.6 & 51.6 & 61.0  & 60.2 & 57.3 & 57.6 & 43.3  & 44.2 & 52.7 & +2.9 \\
\cosine & 70.4 & 59.2 & \sota{52.4} & 60.7 & 57.4 & 56.6 & 54.7 & 44.4   & 40.1 & 51.3 & +0.2 \\
\mi~\citep{li2026mi} \vv{2026.04} &  71.8 & 60.0 & 51.3 & 61.0 & 60.4 & 57.5 & 56.5 & 45.4   & 43.2 & 52.7 & +2.9 \\
VisionZip \citep{yang2025visionzip} \vv{CVPR'25} & 72.3 & \sota{60.1} & {51.6} & 61.3 & 61.0 & \sota{58.5} & 57.8 & 44.7  & 43.1 & 53.1 & +3.7 \\
FastVID \citep{shen2025fastvid} \vv{NeurIPS'25} & 71.0 & 58.3 & 50.6 & 60.0 & 61.4 & 58.1 & 57.7  & 45.0  & 43.2 & 53.0 & +3.5 \\
FlashVID \citep{fan2026flashvid} \vv{ICLR'26}
& \sota{73.0} &  59.3 & 50.3 & 60.9 & 60.2 & 58.4 & \sota{58.4} & 45.0 & 43.5 & 53.2 & +3.9 \\
\rowcolor{lightpink}
\ech
& 72.4 & \sota{60.1} & {51.6} & \sota{61.4} & \sota{61.8} & 58.1 & 57.9 & \sota{48.3} & \sota{46.8} & \sota{54.5} & \tbd{+6.4} \\
\hline
\rowcolor{lightgray}
\multicolumn{12}{c}{\textit{F80 (5$\times$) / 20\%}} 
\\
\rowcolor{white}
\rand & 71.0 & 61.9 & 51.0 & 61.3 &  59.4 & 57.4 & 57.7 & 45.4  & 43.7 & 53.1 & +3.7 \\
\cosine & 71.5 & 59.5 & 53.3 & 61.5 & 59.0 & 56.7 & 55.9 & 46.1  &  39.8 & 52.0 & +1.6 \\
\mi~\citep{li2026mi}  \vv{2026.04} &  73.1 & 61.1 & 52.7 & 62.3 &  61.6 & 57.9 & 55.6 & 47.2  &  42.6 & 53.1 & +3.7 \\
VisionZip \citep{yang2025visionzip}  \vv{CVPR'25} & \sota{74.2} & 60.0 & 52.1 & 62.1 & 60.0 & 58.2 & 57.4 & 43.1 & 41.6 & 52.5 & +2.5 \\
FastVID \citep{shen2025fastvid}  \vv{NeurIPS'25} & 73.0 & 59.9 & 51.7 & 61.5 & 61.2 & 58.4 & 58.0 & 44.4 & 41.8 & 52.8 & +3.1 \\
FlashVID \citep{fan2026flashvid} \vv{ICLR'26} & \sota{74.2} & 60.8 & 52.2 & 62.4 & 61.4 & \sota{58.6} & \sota{58.9} & 45.0  & 41.9 & 53.4 & +4.3 \\
\rowcolor{lightpink}
\ech
& 73.0 & \sota{62.3} & \sota{53.4} & \sota{62.9} & \sota{62.0} & \sota{58.6} & 57.8 & \sota{48.3} & \sota{47.3} & \sota{55.0} & \tbd{+7.4} \\
\hline
\rowcolor{lightgray}
\multicolumn{12}{c}{\textit{F160 (10$\times$) / 10\%}} \\
\rowcolor{white}
\rand & 70.9 & 60.4 & 52.1 & 61.1 & 59.0 & 57.7 & 55.1 & 45.1 & 43.0 & 52.4 & +2.3 \\
\cosine &  69.2 & 61.8 & 52.7 & 61.2 & 59.2 & 56.9 & 54.8 & 44.6 & 41.0 & 51.7 & +1.0 \\
\mi~\citep{li2026mi}  \vv{2026.04} & 71.2 & 62.4 & 53.4 & 62.4 & 61.6 & 58.2 & 56.4 & 46.0  & 39.3 & 52.5 & +2.5 \\
VisionZip \citep{yang2025visionzip}  \vv{CVPR'25} & 70.7 & 60.1 & \sota{53.9} & 61.6 & \sota{61.8} & \sota{59.6} & 56.8 & 45.1 & 40.2 & 52.7 & +2.9 \\
FastVID \citep{shen2025fastvid}  \vv{NeurIPS'25} & 71.2 & 60.6 & 53.8 & 61.9 & 61.2 & 59.1 & 58.0 & 43.8 & 40.9 & 52.7 & +2.9 \\
FlashVID \citep{fan2026flashvid}  \vv{ICLR'26} & {71.4} & {62.2} & 53.7 &{62.4} & 61.2 & 59.5 & \sota{58.9} & {47.5} & 41.6 & 54.0 & +5.5 \\
\rowcolor{lightpink}
\ech
& \sota{72.7} & \sota{63.2} & 53.3 & \sota{63.1} & 60.6 & 58.2 & 58.0 & \sota{50.6} & \sota{47.9} & \sota{55.6} & \tbd{+8.6} \\
\hline
\rowcolor{lightgray} 
\multicolumn{12}{c}{\textit{F320 (20$\times$) / 5\% \textbf{\say{(ultra)}}}} 
\\
\rowcolor{white}
\rand &
71.2 & 59.9 & 51.8 & 60.0 & 60.1 & 57.9 & 58.1 & 43.9 & 42.2 & 52.4 & +2.3 \\
\mi~\citep{li2026mi} \vv{2026.04} & 69.9 & 61.8 & 51.9 & 61.2 & 61.8 & 57.6 & 53.9 & 46.6 & 39.9 & 51.8 & +1.2 \\
VisionZip \citep{yang2025visionzip} \vv{CVPR'25} & 68.7 & 61.1 & 51.5 & 60.4 & 61.9 & 58.1 & 54.2 & 47.0 & 40.6 & 52.1 & +1.8 \\
FastVID \citep{shen2025fastvid} \vv{NeurIPS'25} & 70.1 & \sota{62.0} & 52.1 & 61.4 & \sota{62.2} & 58.3 & 58.1 & 42.7 & 40.0 & 52.1 & +1.8 \\
FlashVID \citep{fan2026flashvid} \vv{ICLR'26} & 
\sota{72.7} & {61.7} & 53.4 & 62.6 & 60.2 & 58.3 & 58.9 & 44.1 & 38.0 & 52.4 & +2.3 \\
\rowcolor{lightpink}
\ech
& 72.2 & 61.4 & \sota{54.8} & \sota{62.8} & \sota{62.2} & \sota{58.7} & \sota{59.0} &  \sota{51.4} & \sota{46.3} & \sota{55.6} & \tbd{+8.6} \\
\bottomrule
\end{tabular}
}
\end{table}

\subsection{Results}
\label{subsec:exp_results}

\paragraph{LLaVA-OV.} 
As shown in Tab.~\ref{tab:results_ov_32}, LLaVA-OV benefits from increased \#frames (\# denotes number), especially on long video benchmarks. For instance, LVBench consists of extremely long videos with an average duration of  4,101 seconds. As a result, higher sampling rates are necessary for its dense actions and long sequence processing. Under the setting of F96/33.3\%, EchoPrune boosts the scores by 7.6\%$^\uparrow$ on LVBench and 3.3\%$^\uparrow$ on average. When pruning under 256 frames, EchoPrune achieves 105\% performance compared with the baseline. 
On VideoMME-Long and MLVU, EchoPrune consistently improves the performance when more frames are available. 
Notably, two generic methods (VisionZip and MI-Pruner) maintain good performance on short videos in VideoMME-Short and Egoschema, while failing to identify pivot tokens in long videos (LongVideoBench, MLVU and LVBench). 
In the ultra compression case, we keep merely 10\% tokens (56,448$^\downarrow$) from 320 frames (62,720 tokens). Under this setting, MI-Pruner, VisionZip and FastVID fail to improve the overall performance, and FastVID even decreases the average score by 7.5\%$^\downarrow$.  
Compared with other pruning methods, EchoPrune achieves SOTA performance in most datasets while bypassing the inefficient segment-and-merge paradigm. Efficiency analysis is reported in Tab.~\ref{tab:eff_llava}.

\paragraph{Qwen2.5VL and Qwen3VL.}
Tab.~\ref{tab:results_qwen2.5_16} sets 16 frames as the baseline following FlashVID \citep{fan2026flashvid}, then shows pruning effects under $4{\sim}20\times$ frames. Despite the crossmodal importance, \cosine~alone (implemented by Eqn.~(\ref{eq:rel_cross})) is a noisy importance estimator in token pruning. In short videos, it performs even worse than random pruning. This explains why most of the previous work advocates vision-based pruning, while none of them explore pure crossmodal scores. Yet, we claim that the synergy of crossmodal relevance and visual redundancy is necessary for optimal token allocation.  
Despite halved token budgets compared with Tab.~\ref{tab:results_ov_32}, EchoPrune demonstrates comparable or even higher scores on 320 frames. Tab.~\ref{tab:results_qwen3} shows pruning results on Qwen3VL. We report two scales, 2B and 8B, to prove our general usage. Fig.~\ref{fig:vis_qwen2_5} visualizes the retained slots in a video after our compression. Benefiting from the semantic reconstruction, EchoPrune captures the second "\texttt{Guitar}" in different colors (top), and yields good OCR capacity (bottom). The logo "\texttt{smule}" and context cues "\texttt{FREE}, \texttt{subscribe}" are retained while the white wall background is masked. See more visualization in the appendix \ref{subsec:app_more_vis}.

\begin{table}[h]
\centering
\caption{\textbf{Performance on Qwen3VL-series.} Setting F32 as the baseline, different pruning methods are compared under $5\times$ frame rates and the same token budgets. }
\label{tab:results_qwen3}
\resizebox{0.99\textwidth}{!}{
\begin{tabular}{lcccccccccccc} 
\toprule
\multirow{2}{*}{\textbf{Method}} & \multirow{2}{*}{\textbf{Setting}} & \multicolumn{4}{c}{\textbf{VideoMME}} & \multicolumn{4}{c}{\textbf{VideoMMMU}*} & \textbf{LongVideo} & \multirow{2}{*}{\textbf{MLVU}} \\ 
 & & Short & Medium & Long & Overall & Per.& Comp. & Adap. & Overall & \textbf{Bench} & \\
\hline
\rowcolor{lightgray}
Qwen3VL-8B & F32 & 75.0 & 59.8 & 56.6 & 63.8 & 73.0 & 64.0  & 35.7 & 57.6 & 58.1 & 46.2 \\
\mi~\citep{li2026mi} \vv{2026.04} & \multirow{4}{*}{\textit{F160/20\%}} & \sota{77.4} & 66.8 & \sota{57.1} & \sota{67.1} &  67.7 & 55.0  & 28.3 & 50.3 & 60.7 & 53.5 \\
VisionZip \citep{yang2025visionzip} \vv{CVPR'25} & & 77.2 & 65.9 & 56.5 & 66.5 & 68.1 & 56.2 & 28.4 & 50.9 & 60.8 & 53.1 \\
FlashVID \citep{fan2026flashvid} \vv{ICLR'26} & & 77.0 & 65.7 & 55.1 & 65.9 & 71.7 & 58.7  & 36.3 & 55.6 & \sota{61.6} & \sota{53.8} \\
\rowcolor{lightpink}
\ech & & 75.2 & \sota{67.8} & \sota{57.1} & {66.7} & \sota{74.3} & \sota{61.3}  & \sota{38.3} & \sota{58.0} & 59.9 & \sota{53.8} \\
\hline
\rowcolor{lightgray}
Qwen3VL-2B & F32 & 68.3 & 52.9 & 48.6 & 56.6 & 55.0 & 43.0  & 26.0  & 41.3 & 51.6 & 42.3 \\
MI-Pruner~\citep{li2026mi} \vv{2026.04} & \multirow{4}{*}{\textit{F160/20\%}} & 72.6 & 59.0 & \sota{47.1} & 59.6 &  48.7 & 35.3  & 26.7  & 36.9 & \sota{54.8} & 47.4 \\
VisionZip \citep{yang2025visionzip} \vv{CVPR'25} & & 72.8 & 59.9 & 47.0 & 59.9 & 48.6 & 36.6 & 27.1 & 37.4 & 54.3 & 47.1 \\
FlashVID \citep{fan2026flashvid} \vv{ICLR'26} & & 72.1 & 57.0 & 47.0 & 58.7 & 45.0 & 37.3  & 22.0 & 34.8 & 52.6 & 45.3 \\
\rowcolor{lightpink}
\ech & & \sota{73.1} & \sota{59.2} & 46.9 & \sota{59.8} & \sota{56.7} & \sota{40.7}  & \sota{27.7} & \sota{41.7} & \sota{54.8} & \sota{47.6} \\
\bottomrule
\multicolumn{12}{l}{* Per./Comp./Adap. of VideoMMMU represent Perception/Comprehension/Adaptation splits, respectively.}
\end{tabular}
}
\end{table}

\begin{table}[h]
\centering
\vspace{-0.3cm}
\caption{\textbf{Efficiency on LLaVA-OV-7B.} We report the compression time, prefilling time and Time-To-First-Token (TTFT) in milliseconds. Our speedup on prefilling and TTFT is annotated in (N$\times$).}
\label{tab:eff_llava}
\resizebox{0.99\textwidth}{!}{
\begin{tabular}{l|ccccc|ccccc} 
\toprule
\textbf{Method} & \textbf{Settings} & \textbf{Compress.} & \textbf{Prefilling} & \textbf{TTFT} & \textbf{Scores}  & \textbf{Settings} & \textbf{Compress.} & \textbf{Prefilling} & \textbf{TTFT} & \textbf{ Scores} \\
\hline
\rowcolor{lightgray}
LLaVA-OV-7B & F64 & - & 917.2 (${1\times}$) & 1569.4 (${1\times}$) & 58.8 & F160 & - & 2760.8 (${1\times}$) & 4391.6 (${1\times}$) & 58.6 \\
HoliTom \citep{shao2025holitom} \vv{NeurIPS'25} & \multirow{4}{*}{\textit{F64/20\%}} & 280.7 & 523.7 & 1175.9 & 58.4  & \multirow{4}{*}{\textit{F160/20\%}} & 830.4 & 1423.0 & 3053.8 & 60.1 \\
UniComp \citep{yuan2026unicomp} \vv{CVPR'26} &  & 210.1 &  353.5 & 1005.7 & 58.6 &  & 511.2 & 818.8 & 2449.6 & 60.2 \\
FlashVID \citep{fan2026flashvid} \vv{ICLR'26} &  & 92.1 &  258.0 & 910.2 & 58.5 &  & 180.3 & 624.4 & 2255.2 & 60.0 \\
\rowcolor{lightpink}
\ech  &  & 18.2 & 192.9 \tbd{($\mathbf{4.8\times}$)} & 845.1 \tbd{(${1.9\times}$)} & 58.7 & & 42.8 & 491.3 \tbd{($\mathbf{5.6\times}$)} & 2122.1 \tbd{(${2.1\times}$)} & 60.4 \\
\bottomrule
\end{tabular}
}
\vspace{-0.5cm}
\end{table}

\paragraph{Efficiency analysis.} Tab.~\ref{tab:eff_llava} assesses the efficiency of LLaVA-OV-7B. We report average values of VideoMME after warmups. The Time-To-First-Token (TTFT) consists of ViT encoding and prefilling (compression and LLM forward). Since all pruning methods share the same ViT, we report the compression costs, overall prefilling time and TTFT in Tab.~\ref{tab:eff_llava}. Due to the limited capacity, increasing frames to F160 doesn't boost the baseline performance (58.8 to 58.6). In comparison, our pruning reduces the \#tokens to the same as the default F32 setting, improving the score to 60.4.
On these two settings, EchoPrune accelerates around $5\times^\uparrow$ on prefilling time and $2\times^\uparrow$ on TTFT.

\begin{figure}[b]
\centering
\vspace{-0.2cm}
\begin{minipage}[h]{\textwidth}
    \centering
    \includegraphics[width=\linewidth]{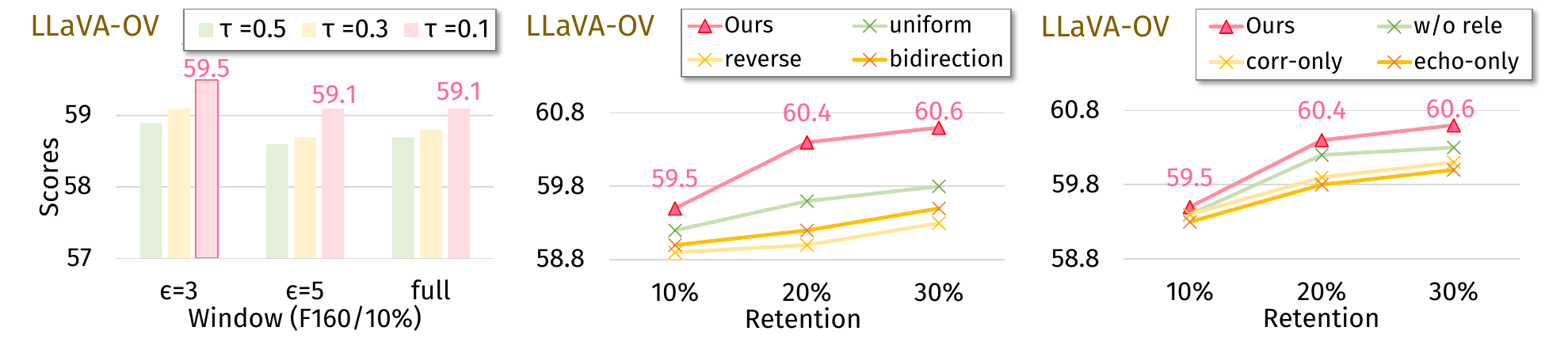}
\end{minipage}
\vspace{0.2cm}
\begin{minipage}[h]{\textwidth}
    \centering
\includegraphics[width=\linewidth]{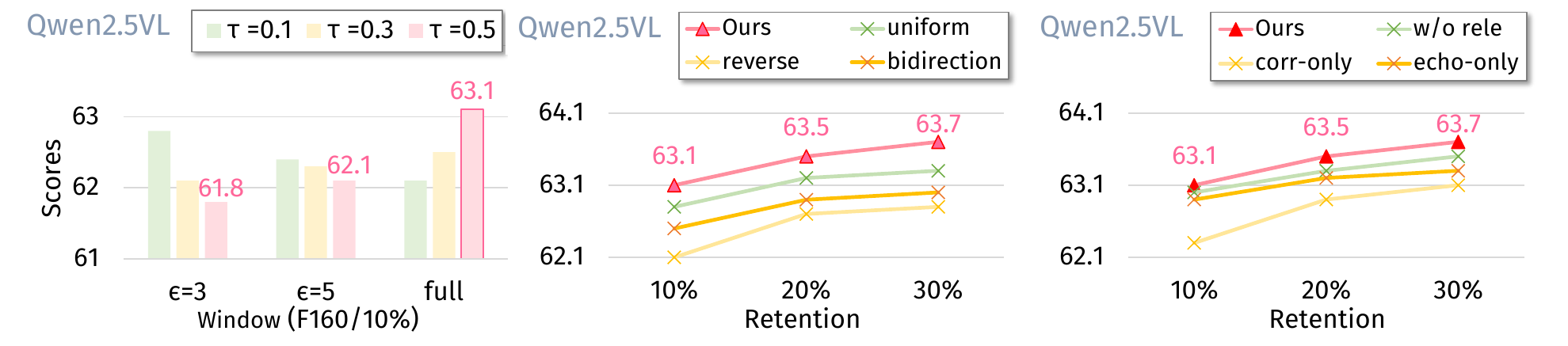}
\end{minipage}
\vspace{-0.2cm}
\caption{\textbf{Ablation study.} \textbf{Left:} LLaVA-OV performs best on $0.1\tau$ and $\epsilon{=}3$, while Qwen2.5VL suits $0.5\tau$ and full-frame matching. \textbf{Middle:} Our global Top-K and causal $\delta$ is better than uniform, reverse and bidirectional designs. \textbf{Right:} The synergy yields SOTA results compared with single types.}
\label{fig:ablation}
\end{figure}

\subsection{Ablation Study}
\label{subsec:exp_ablation}

Fig.~\ref{fig:ablation} shows ablation study on VideoMME (F160) for LLaVA-OV-7B (top) and Qwen2.5VL-7B (bottom). See discussions on Qwen3VL in App.~\ref{subsec:app_more_exp}.
Fig.~\ref{fig:ablation}(left) contrasts various configurations of $\tau~\text{and}~\epsilon$. We select temperature $\tau{=}0.1$ and window size $\epsilon{=}3$ for LLaVA-OV-7B, and holistic matching with $0.5\tau$ for QwenVL-series. Fig.~\ref{fig:ablation}(middle) illustrates the effectiveness of our global Top-K ranking and causal reconstruction. Specifically, the \textit{uniform} setting replaces the global Top-K with per-frame Top-K under a uniform frame budget, which leads to diluted importance. The \textit{reverse} defines $\delta_\mathrm{echo}$ as the backward error (from frame $k{+}1$), and \textit{bidirection} averages over forward (from $k{-}1$) and backward (from $k{+}1$) errors. Both yield inferior performance compared to our forward strategy, likely due to our preservation of temporal causality.
Finally, our synergy design consisting of correspondence and echo matching is better than the one-type pattern, as reported in Fig.~\ref{fig:ablation}(right). The echo matching only (\textit{echo-only}) works better than correspondence matching only (\textit{corr-only}) in Qwen2.5VL for its stronger ViT encoder. And \textit{w/o rele} removes the crossmodal relevance term (Eqn.~(\ref{eq:rel_cross})) for purely redundancy-guided ranking, which undermines the overall performance. See more discussions about the range of frame matching in App.~\ref{subsec:app_more_exp}.

\section{Related Work}
\label{sec:related}
\vspace{-2mm}

\textbf{VideoLLMs.}
The advanced multimodal language models have a strong video understanding capacity \citep{wang2025internvl3,Qwen3VL}. 
Following the \texttt{Enc-MLP-Dec} architecture \citep{liu2024improved}, LLaVA-OV \citep{li2024llavaov} improves the MLLM performance in single-image, multi-image and video scenarios. The latest Qwen2.5VL \citep{qwen2.5-VL} and Qwen3VL \citep{Qwen3VL} further support adaptive resolution with stronger perception and comprehension performance.
Despite the progress in VideoLLMs, their substantial GPU memory consumption and high inference latency pose critical bottlenecks for real-world deployment \citep{luo2024gamma,liao2026resadapt}. Consequently, a growing body of research has diverted its focus toward efficient inference techniques \citep{shen2025fastvid,tao2025dycoke}, encompassing both training-based and training-free paradigms.

\textbf{Training-based token compression.}
Within the realm of training-based methods, the series of Mixture-of-Depths (MoD) \citep{luo2024gamma,zhang2025p} employ a learnable router in LLMs to identify essential vision tokens while bypassing redundant ones. In parallel, low-resolution (LR) visual inputs naturally require fewer tokens and suffice for addressing coarse-grained queries. Recent studies \citep{yang2025visionthink,liao2026resadapt} advocate resolution-on-demand, which frames the policy of \textit{whether to request high resolution} through the lens of reinforcement learning (RL). Formulating reconstruction error as token uniqueness, AutoGaze \citep{shi2026attend} incorporates VideoMAE \citep{tong2022videomae} to perform autoregressive reconstruction before ViTs. Likewise, STTS \citep{zhang2026unified} trains a scorer in ViTs with a reconstruction loss. Yet, the pruning on the visual encoder side needs extra tuning for multimodal alignment. 

\textbf{Training-free token compression.}
The training-free token reduction techniques are notable for their versatility and efficiency. 
The \textit{generic methods} are applicable to both image and video pruning. For instance, VisionZip \citep{yang2025visionzip} and so on \citep{zhang2024sparsevlm,zhang2025beyond} leverage attention matrices to identify salient tokens, and other methods analyze token similarities and coverages under a subset selection framework \citep{zhang2025cdpruner,alvar2025divprune,li2026mi}.
A conventional pipeline adopted by many \textit{video-specific} methods \citep{shen2025fastvid,shao2025holitom,yuan2026unicomp} is characterized as segment-and-merging. Alternatively, FlashVID \citep{fan2026flashvid} and ForestPrune \citep{ju2026forestprune} organize tokens into tree-based structures for spatiotemporal modeling.
Recent image pruning approaches \citep{li2026resprune,ma2026apet} leverage the \textit{reconstruction error} as an importance measure, since unique tokens are not predictable \citep{li2026resprune,wang2026pixelprune}. This principle is also extended to video compression \citep{shi2026attend,zhang2026unified}.
Yet, reconstruction-based pruning in a training-free manner remains relatively underexplored.
To bridge the gap, we introduce temporal echo reconstruction and query-aware retention. Please refer to App.~\ref{subsec:app_related_videollm} for a wider discussion.

\section{Conclusion}
\label{sec:conclusion}

In this paper, we propose a training-free video pruning method for efficient VideoLLMs. Motivated by maximum marginal relevance, \ech~employs query-guided \textit{crossmodal relevance} and \textit{temporal reconstruction error} for explicit spatiotemporal modeling. Working as a plug-in module, our method supports once-for-all compression before LLMs, bypassing segment and merging operations.
Extensive experiments on LLaVA-OV, Qwen2.5VL and Qwen3VL across six benchmarks demonstrate our SOTA performance and efficiency. Under a fixed token budget, our method expands the processable frame rate to 20$\times$ with an average 7\%$^\uparrow$ performance gain. Furthermore, \ech~accelerates the prefilling by 5.6$\times$ and TTFT by 2.1$\times$ on LLaVA-OV, which enables fine-grained long-term video understanding and low-latency responses in computation-limited scenarios. 

\textbf{Limitations and future work.} Despite the benefits, there exist three drawbacks in our method. \textit{(i)} For simplicity, we prune tokens once in the projection layer, without ViT acceleration strategies \citep{yu2026visiontrim}.
\textit{(ii)} Our method retains the original indices for Rotary Position Embedding (RoPE). We believe that enhanced RoPE will boost the performance further \citep{gong2025echoingpixels}. \textit{(iii)} We only consider the vision modality, regardless of auditory streams. The audio-guided video pruning \citep{tao2025omnizip} for Omnimodal Large Language Models becomes future work.

\begin{ack}

This research received funding from the Flemish Government (AI Research Program) and the Research Foundation
Flanders (FWO) through project number G0G2921N.
We acknowledge EuroHPC JU for awarding access to the EuroHPC supercomputer LEONARDO, hosted by CINECA (Italy) and the LEONARDO consortium. We thank Han Zhou for helpful discussions.

\end{ack}

\bibliographystyle{plainnat}
\bibliography{ref}

\clearpage

\appendix

\begin{center}
    \rule{0.99\textwidth}{1pt} \\[0.2em]
    \LARGE \textbf{Appendix of EchoPrune: Interpreting Redundancy as Temporal Echoes for Efficient VideoLLMs} \\
    \rule{0.99\textwidth}{1pt}
\end{center}

\section{Extended Related Work}
\label{sec:app_related}

\subsection{Efficient VideoLLMs}
\label{subsec:app_related_videollm}

We extend the technical background to demonstrate why training-free token pruning is a promising avenue for achieving efficient VideoLLMs.
As mentioned in Sec.~\ref{sec:intro}, the \textbf{Mixture-of-Depth} (MoD) mechanism operates on each LLM layer to select essential vision tokens for processing while skipping redundant ones. Typically, their MoD modules \citep{luo2024gamma,zhang2025p} learn to assess the token importance with limited training data. A recent work, p-MoD \citep{zhang2025p} proposes a progressive ratio decay from shallow to deep layers. 
Back to the raw representation, \textbf{codec-aware} methods like OneVision-Encoder\citep{tang2026onevision} and CoPE-VideoLM \citep{sarkar2026cope}, leverage the codec information (e.g., motion vectors and residual signals) from video data to encode what moves and what changes between frames, thus preserving the sparsity of the original stream. CoPE-VideoLM \citep{sarkar2026cope} encodes the motion vectors and residuals (by $\Delta$-encoder), then concatenates them with the keyframe tokens (from vision encoder) as the visual inputs, which supports dense temporal coverage at a fraction of the typical token budgets and runtime.
Another technique is \textbf{speculative decoding}, which leverages a small model to generate a bunch of tokens and a large model to do one-time validation. One of its well-known applications in LLMs is the EAGLE-series \citep{li2024eagle,li2024eagle2,li2025eagle3}.
In VideoLLMs, SpecVLM \citep{ji2025specvlm} proposes vision-aware drafting, including attention-guided Top-P retention and spatial uniform sampling.
The following work LVSpec \citep{ji2026see} introduces vision-aware verifying, and ParallelVLM \citep{kong2026parallelvlm} employs parallel drafting and verifying.
Beyond token operations, a distinct line of research focuses on \textbf{keyframe selection}, which replaces the uniform downsampling initially in VideoLLMs with adaptive algorithms. For instance, AKS \citep{tang2025adaptive} downsamples the raw video into 1 frame per second and employs BLIP-ITM \citep{li2022blip} to conduct adaptive keyframe sampling until the frame budgets. 
Similarly, GIFT \citep{ma2026gift} starts from uniformly sampled 128 frames and utilizes SigLip \citep{zhai2023sigmoid} to filter out keyframes by irreplaceability and budget-aware refinement. 
These frame selectors work as a pre-filtering module, which feed selected frames into VideoLLMs with the help of external Vision Language Models.
Despite the efficiency benefits of the above methods, their implementation and application are less flexible than training-free token pruning in practice. We study token pruning for its superior versatility and minimal intrusiveness of the pretrained model.

\begin{figure}[b]
\centering
\includegraphics[width=\linewidth]{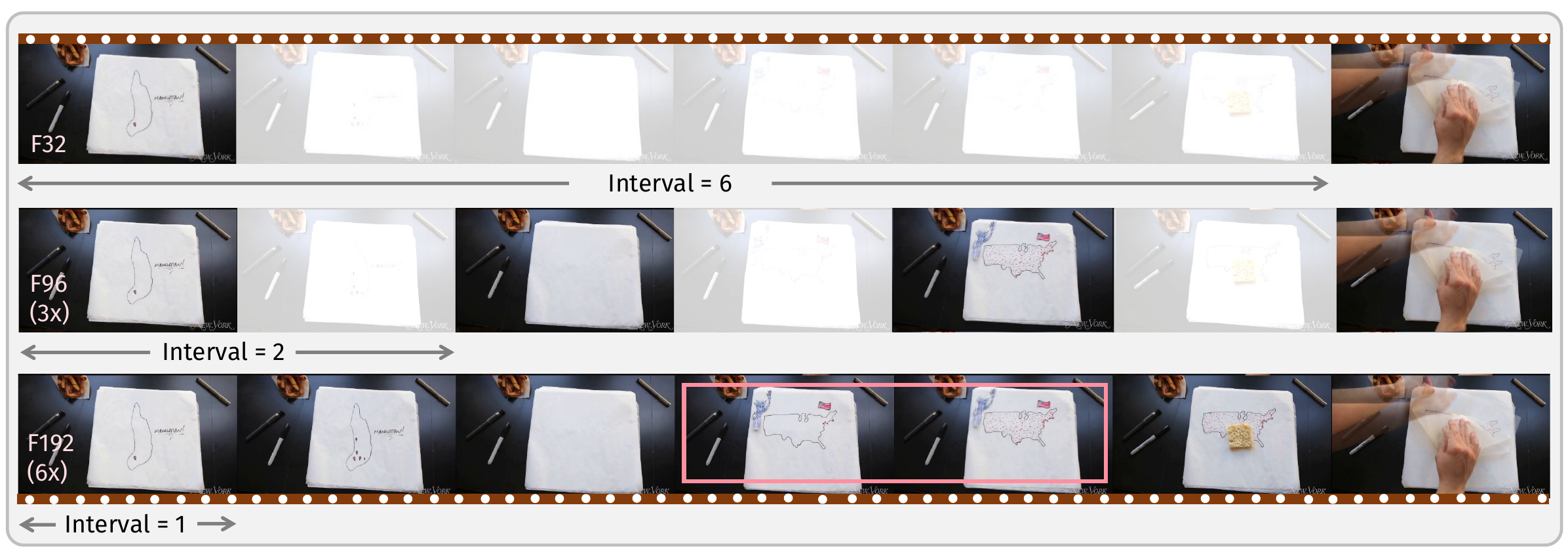} 
\caption{
\textbf{Temporal resolution under different sampling rates.} 
The light-colored frames represent dropped clips under decreasing temporal resolution.
Due to the large sampling interval, the pivot frames{\setlength{\fboxrule}{1.5pt}\setlength{\fboxsep}{2.5pt}\fcolorbox{pink}{white}{}} are fully skipped in the default F32 {(the top row)}. 
}
\label{fig:sampling}
\end{figure}

\subsection{Rethinking Temporal Sampling and Aliasing Dilemma}
\label{subsec:app_related_sample}

The long-video sampling issue in VideoLLM benchmarks was first noticed by LSDBench \citep{qu2025does}. It introduces the concept of \textit{necessary sampling density}, defined as the minimum sampling density required to accurately answer a given question.
Here, we discuss the undersampling bias mentioned in the main paper by reviewing the sampling interval ($I$) and temporal resolution ($\mathrm{TR}$) from sampling theory.
Specifically,  the sampling interval refers to the stride between two consecutively sampled frames from a raw video sequence. Accordingly, temporal resolution characterizes the density of visual information (represented by the frame rate) captured per unit of time, which is the inverse of the sampling interval ($\mathrm{TR}{=}\frac{1}{I}$). 
When uniformly sampling $M_F$ frames from $M$ frames, the interval $I$ and the temporal resolution $\mathrm{TR}$ are written as:
\begin{align}
    I&=\frac{M}{M_F};
    \label{eq:interval}\\
    \mathrm{TR}&=\frac{1}{I}=\frac{M_F}{M}.
    \label{eq:tr}
\end{align}

\begin{wraptable}{r}{0.5\textwidth} 
    \centering
    \vspace{-0.2cm}
    \caption{Temporal sampling$_{M{=}192}$.}
    \vspace{-0.2cm}
    \label{tab:t_sampl}
    \begin{tabular}{lccc}
    \toprule
     & F32 & F96 & F192 \\
    \hline
    Interval ($I$) & 6 & 2 & 1 \\
    Temporal resol. ($\mathrm{TR}$) & $\frac{1}{6}$ & $\frac{1}{2}$ & 1 \\
    \bottomrule
    \vspace{-0.5cm}
    \end{tabular}
\end{wraptable}
Fig.~\ref{fig:sampling} treats 192 frames as the fully sampled baseline ($M{=}192$), \textit{i.e.}, F192 has a sampling interval $I{=}1$ and temporal resolution $\mathrm{TR}{=}1$. 
As mentioned in Sec.~\ref{sec:intro}, VideoLLMs uniformly sample several frames (denoted as $M_F$) from the raw video for the limited model capacity.
To reduce $M_F$ for faster inference, the interval increases to 2 with a halved temporal resolution ($\mathrm{TR}{=}\frac{1}{2}$), and the interval reaches 6 in a sparsely downsampled case F32 ($\mathrm{TR}{=}\frac{1}{6}$). We summarize three groups of $I$ and $\mathrm{TR}$ in Tab.~\ref{tab:t_sampl}.
The critical frames required to answer the query are highlighted{\setlength{\fboxrule}{1.5pt}\setlength{\fboxsep}{2.5pt}\fcolorbox{pink}{white}{}} in Fig.~\ref{fig:sampling}. 
Notably, F32 fails to capture these pivot moments due to its large sampling interval. As a result, the LLM is forced to generate responses based solely on pretrained priors rather than the actual visual evidence.

According to Nyquist sampling theory \citep{nyquist1928certain}, \textit{a continuous signal can only be perfectly reconstructed if the sampling rate is at least twice the highest frequency component present in the signal}. Therefore, aliasing occurs in F32 (Fig.~\ref{fig:sampling}, top), where the frame rate drops below the critical threshold, making lossless reconstruction mathematically impossible. Despite the reduced computation by processing fewer frames, the signal transmission in such a sparse-sampling scenario is fundamentally lossy. This prompts us to rethink the previous setting \citep{shen2025fastvid,shao2025holitom} of pruning on F32 and comparing TTFT and FLOPs to show efficiency, since they overlook the unfairness caused by a "blinded LLM". To address it, our benchmark maintains the same token budget as the original F32 but scales the visible frames. From the view of signal processing, we increase the number of visible frames from $M_F$ to $\gamma M_F (\gamma>1)$  in order to satisfy the reconstruction threshold in Nyquist sampling theory. Then, we conduct pruning by temporal reconstruction to keep the high-frequency components.

\begin{figure}[h]
\centering
\includegraphics[width=\linewidth]{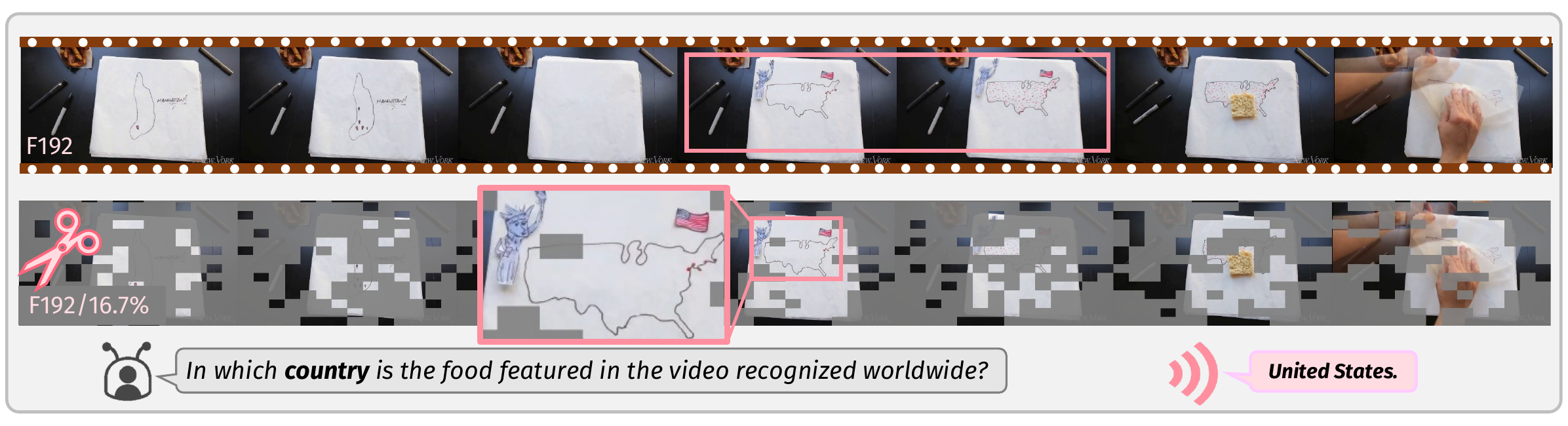} 
\caption{\textbf{Interpretability of EchoPrune (LLaVA-OV-7B).} To avoid undersampling, we first allow sufficient visible frames and then prune tokens (F192/16.7\%) to reach the budgets of full-frame F32.
}
\label{fig:vis_llava}
\end{figure}

As shown in Fig.~\ref{fig:vis_llava}, by preserving sufficient temporal resolution before pruning, we ensure that VideoLLMs "see" the essential content and then decide which tokens to proceed in decoding. 
Benefiting from the temporal echo mechanism, EchoPrune selectively preserves the first pivot frame with high integrity, while the majority of the second pivot frame is pruned since it's reconstructable from the first frame. This behavior effectively demonstrates the interpretability of our method in identifying and removing temporal redundancy.

\section{Extended Experiments}
\label{sec:app_exp}

\subsection{Implementation Details}
\label{subsec:app_imple}

\paragraph{Efficiency analysis.} The TTFT consists of ViT encoding, compression and LLM forward. 
In Tab.~\ref{tab:eff_llava}, we report the compression time, prefilling time (compression and LLM forward) and TTFT on a single A100 GPU by averaging over 100 samples in VideoMME. For each sample, we conduct 1 warmup and 3 runs to get the average time. 

\subsection{Dataset Description}
\label{subsec:app_dataset}

\textbf{VideoMME} \citep{fu2025video}. VideoMME comprises 900 videos spanning 6 diverse domains and 30 subcategories. The video durations range from 11 seconds to one hour, with an average duration of 1,018 seconds. Each video is accompanied by high-quality human annotations, including \colorbox{beige}{2700} multiple-choice question-answer pairs.

\textbf{EgoSchema} \citep{mangalam2023egoschema}. EgoSchema includes \colorbox{beige}{5031} multiple-choice questions derived from 250 hours of egocentric video. It emphasizes long-range temporal reasoning and poses substantial challenges for both fine-grained spatial perception and sustained temporal coherence. The subset is a reduced version of the full benchmark, with \colorbox{beige}{500} questions.

\textbf{LongVideoBench} \citep{wu2024longvideobench}. LongVideoBench consists of 3,763 videos from 8 seconds to 1 hour, and 6,678 human-annotated multiple-choice questions. The wide variation in video length and content makes it particularly challenging to evaluate temporal modeling, contextual understanding, and crossmodal alignment in video-language models. We tested on the validation set with \colorbox{beige}{1337} questions.

\textbf{MLVU} \citep{zhou2025mlvu}.
MLVU comprises 3,102 multiple-choice questions spanning 9 long video understanding tasks. The videos range from 3 minutes to 2 hours, requiring reasoning over narrative structure, temporal order, and event retrieval. This benchmark jointly evaluates fine-grained spatial perception and long-range temporal reasoning capabilities. The test set includes \colorbox{beige}{502} question-answer pairs.

\textbf{VideoMMMU} \citep{hu2025video}. VideoMMMU evaluates the professional reasoning capacity in academic and scientific areas. The 300 expert-level videos have an average duration of 506 seconds, categorized into adaptation, comprehension and perception.  
It involves \colorbox{beige}{900} multiple-choice questions. 

\textbf{LVBench} \citep{wang2025lvbench}. 
LVBench is designed to challenge multimodal models' long-term memory and extended comprehension capabilities.
Targeted at extremely long video understanding, 
LVBench contains 103 videos with an average duration of 4,101 seconds. To meet the demands of real-world applications, LVBench consists of \colorbox{beige}{1549} manually annotated multiple-choice questions, spanning over six QA tasks.

\begin{figure}[h]
\centering
\begin{minipage}{0.61\textwidth}
    \centering
    \includegraphics[width=\textwidth]{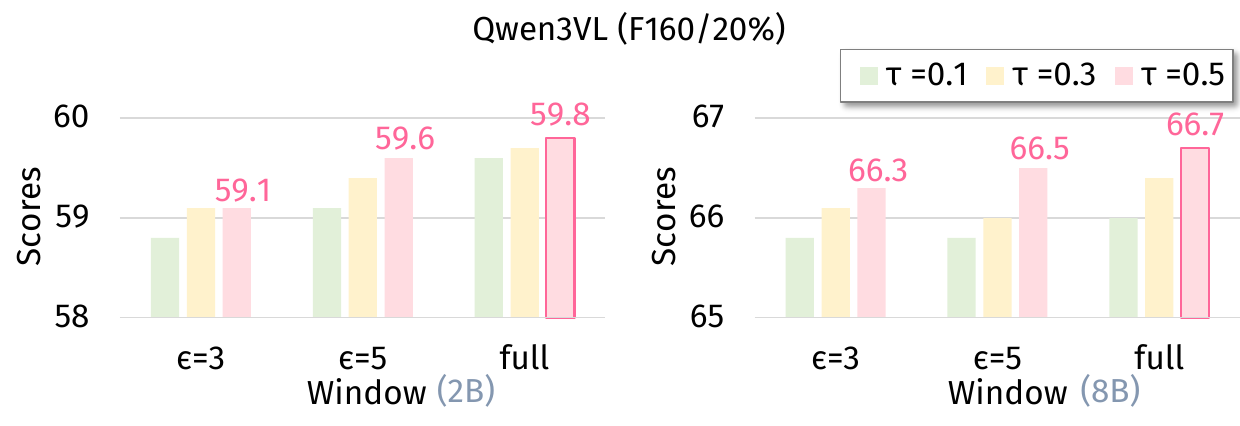}
    \captionof{figure}{Ablation study on Qwen3VL.}
    \label{fig:ablation_qwen3}
\end{minipage}
\begin{minipage}{0.36\textwidth}
    \centering
    \includegraphics[width=\textwidth]{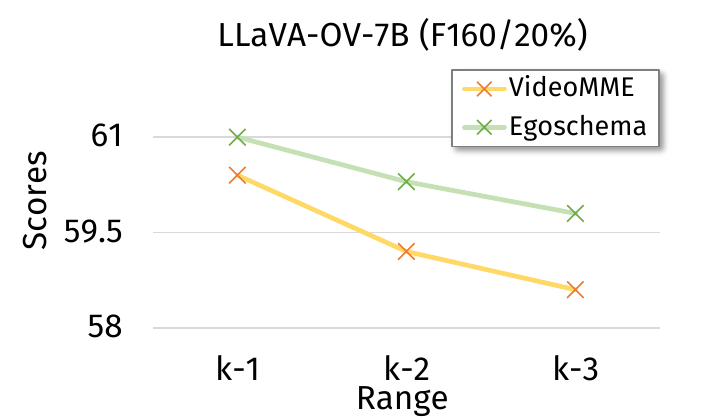}
    \captionof{figure}{Various matching ranges.}
    \label{fig:markov}
\end{minipage}
\end{figure}

\subsection{Extended Evaluation}
\label{subsec:app_more_exp}

\paragraph{Ablation on Qwen3VL.} 
Instead of per-dataset tuning, we adopt $\tau{=}0.5$ and full-frame matching for QwenVL-series. 
We show the corresponding ablation study on Qwen2.5VL in Fig.~\ref{fig:ablation}, and extend the comparison on
Qwen3VL-2B and 8B in Fig.~\ref{fig:ablation_qwen3} to justify our choice. In accordance with Qwen2.5VL, Qwen3VL exhibits SOTA performance on the same configuration ($\tau{=}0.5$, full-frame). These empirical results demonstrate that our approach is user-friendly in terms of parameter configuration, requiring minimal effort for optimization.

\paragraph{Ablation on frame matching.} 
We adopt the adjacent frame matching in Sec.~\ref{subsec:method_recon}, which enforces a localized temporal consistency constraint of the reconstruction error $\delta(\tilde{\mathbf{v}}^{k}_i,\tilde{\mathcal{V}}^{k-1})$.
From the viewpoint of efficiency, we aim to avoid laborious iteration over all history frames from 1 to $k{-}1$. 
In Fig.~\ref{fig:markov}, we provide empirical justification by extending the frame range to $k{-}2$ and $k{-}3$, including the nearest two frames and three frames, respectively. The range of $k{-}1$ involves tokens from the last frame (adopted by us), while the range of $k{-}2$ includes the last two frames with doubled candidate tokens, and the range of $k{-}3$ includes tripled.
Empirically, the increased matching range undermines performance due to the diluted temporal correspondence, which justifies our efficient choice.

\begin{figure}[h]
\centering
\begin{minipage}{0.62\textwidth}
    \centering
    \includegraphics[width=\textwidth]{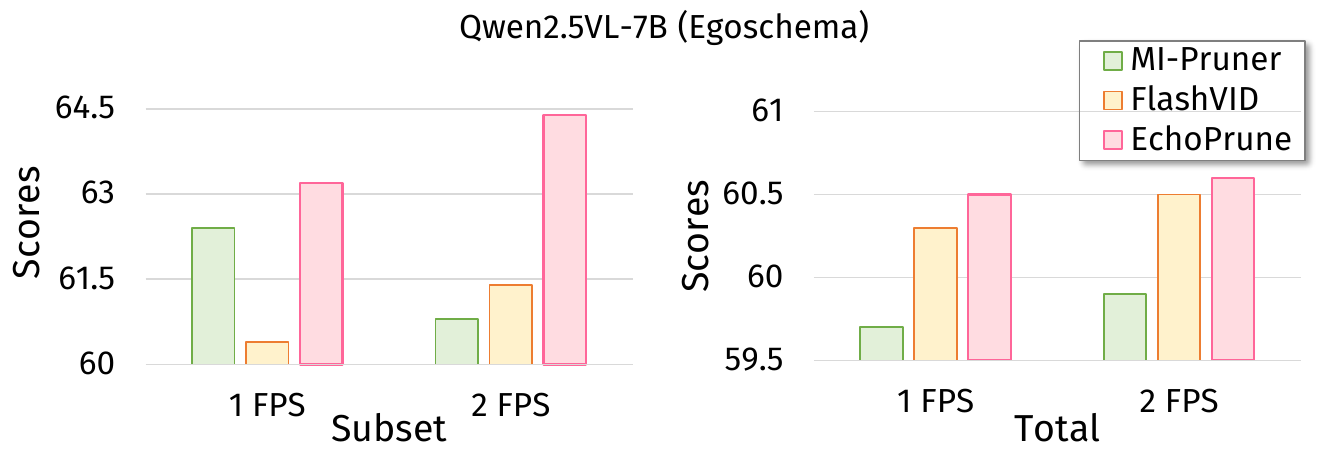}
    \captionof{figure}{Performance under various FPS.}
    \label{fig:fps}
\end{minipage}
\begin{minipage}{0.37\textwidth}
    \centering
    \includegraphics[width=\textwidth]{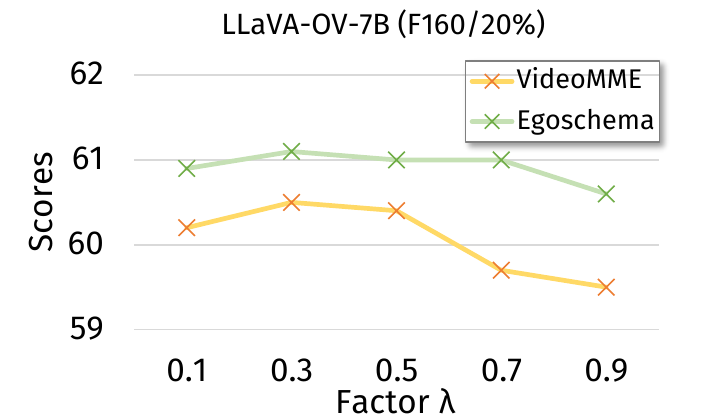}
    \caption{Various $\lambda$ in MMR.}
    \label{fig:lambda}
\end{minipage}
\end{figure}

\paragraph{Pruning under various FPS.} In the main paper, we compare EchoPrune with SOTA methods under varying \#frames. Here, we report Qwen2.5VL-7B performance under various frames per second (FPS). As shown in Fig.~\ref{fig:fps}, we keep 20\% tokens for both subset and total settings on Egoschema. EchoPrune consistently achieves SOTA performance on 1 FPS and 2 FPS, which demonstrates our robustness across various configurations.

\paragraph{More model scales.} 
Tab.~\ref{tab:results_qwen2.5_3b} reports the pruning effects on 7B and 3B scales compared with three methods: random sampling, MI-Pruner \citep{li2026mi} (generic pruning) and FlashVID \citep{fan2026flashvid} (video pruning). 
In the official repository, FlashVID sets the 20$_\text{th}$ layer as the LLM pruning layer in Qwen2.5VL-7B.
Since Qwen2.5VL-3B \citep{qwen2.5-VL} has 36 LLM layers, we implement  25$_\text{th}$ layer as the LLM pruning layer. The settings of MI-Pruner and EchoPrune are the same for 3B and 7B models.
We conduct experiments on \textit{F160/20\%} settings. EchoPrune improves the average score by 5.7\%$^\uparrow$ for the 7B model and 6.0\%$^\uparrow$ for 3B, which shows our robustness.

\begin{table}[h]
\centering
\caption{\textbf{Performance on Qwen2.5VL-7B and -3B.} Two baselines process 32 frames, while pruning methods are compared under 160 frames and 20\% retained. 
}
\label{tab:results_qwen2.5_3b}
\resizebox{0.99\textwidth}{!}{
\begin{tabular}{lccccccccc|cc}
\hline
\multirow{2}{*}{\textbf{Method}} & \multicolumn{4}{c}{\textbf{VideoMME}} & \multicolumn{2}{c}{\textbf{EgoSchema}} & \textbf{LongVideo} & \multirow{2}{*}{\textbf{MLVU}} & \textbf{Video} & \multicolumn{2}{c}{\textbf{Average}} \\ 
 & Short & Medium & Long & Overall & Subset & Total & \textbf{Bench} & & \textbf{MMMU} & Scores & \% \\
\hline
\rowcolor{lightgray}
\multicolumn{12}{c}{\textit{32 frames (1$\times$) / keep 100\%}} \\
\rowcolor{white}
Qwen2.5VL-7B & 71.2 & 58.8 & 51.0 & 60.3 & 60.6 & 58.1 & 58.9 & 44.6 & 48.7 & 54.1 & 100 \\
\hline
\rowcolor{lightgray}
\multicolumn{12}{c}{\textit{160 frames (5$\times$) / keep 20\%}} \\
\rowcolor{white}
\rand 
&  74.4 & 63.9 & 52.3 & 63.6 & 61.0 & 59.0 & 60.5 & 49.7 & 42.3 & 55.0 & +1.7 \\
MI-Pruner \citep{li2026mi} \vv{2026.04}
&  74.6 & 65.2 & 54.2 & 64.7 & 63.0 & 59.8 & 59.3 & 49.6 & 39.3 & 54.5 & +0.7 \\
FlashVID \citep{fan2026flashvid} \vv{ICLR'26}
&  75.3 & 64.8 & 55.1 & 65.1  & 61.0 & 60.1 & 61.6 & 48.6 & 43.2 & 55.7 & +3.0 \\
\rowcolor{lightpink}
\ech
&  76.1 & 65.2 & 55.0 & 65.4 & 61.2 & 59.9 & 61.1 & 51.4 & 48.4 & 57.2 & \tbd{+5.7} \\
\hline
\rowcolor{lightgray}
\multicolumn{12}{c}{\textit{32 frames (1$\times$) / keep 100\%}} \\
\rowcolor{white}
Qwen2.5VL-3B & 68.6 & 56.1 & 47.9 & 57.5  & 53.8 & 54.3 & 54.6 & 43.5 & 41.1 & 50.2 & 100 \\
\hline
\rowcolor{lightgray}
\multicolumn{12}{c}{\textit{160 frames (5$\times$) / keep 20\%}} \\
\rowcolor{white}
\rand 
&  70.3 & 60.1 & 48.3 & 59.6 & 55.2 & 55.0 & 55.1 & 48.0 & 40.1 & 51.6 & +2.8 \\
MI-Pruner \citep{li2026mi} \vv{2026.04}
&  71.0 & 59.3 & 49.0 & 59.8 & 55.8 & 55.6 & 56.4 & 48.4 & 40.2 & 52.1 & +3.8 \\
FlashVID \citep{fan2026flashvid} \vv{ICLR'26}
&  72.3 & 60.7 & 49.8 & 60.9  & 56.2 & 56.2 & 55.9  & 48.9 & 40.7 & 52.5 & +4.6 \\
\rowcolor{lightpink}
\ech
&  72.5 & 60.9 & 50.0 & 61.1 & 55.4 & 55.7 & 55.9 & 51.8 & 43.1 & 53.5 & \tbd{+6.0} \\
\hline
\end{tabular}
}
\end{table}

\paragraph{MMR variants.} 
In the main paper, we adopt a principled format of MMR to avoid introducing more hyperparameters, since both relevance and redundancy terms are derived by an inner product measure. Here, we recall Eqn.~(\ref{eq:score}):
\begin{align}
S(\tilde{\mathbf{v}}^{k}_i,\tilde{\mathcal{V}},\tilde{\mathcal{T}}) &= r(\tilde{\mathbf{v}}^{k}_i,\tilde{\mathcal{T}})- \delta(\tilde{\mathbf{v}}^{k}_i,\tilde{\mathcal{V}}^{k-1}).
\end{align}
and report more results when the relevance term is emphasized ($\lambda>0.5$) or vice versa ($\lambda<0.5$):
\begin{align}
S(\tilde{\mathbf{v}}^{k}_i,\tilde{\mathcal{V}},\tilde{\mathcal{T}}) &= \lambda \cdot r(\tilde{\mathbf{v}}^{k}_i,\tilde{\mathcal{T}})- (1-\lambda) \cdot \delta(\tilde{\mathbf{v}}^{k}_i,\tilde{\mathcal{V}}^{k-1}).
\end{align}
We conduct experiments on VideoMME and Egoschema under the setting of {F160/20\%}.
As shown in Fig.~\ref{fig:lambda}, various choices of $\lambda$ don't exhibit a clear benefit. On the contrary, relying more on the relevance (\textit{e.g.}, $\lambda=0.9$) term is harmful. Therefore, we keep the principled format in Eqn.~(\ref{eq:score}), which can be understood as $\lambda{=}0.5$.  

\paragraph{Image pruning.} 
By design, the video and image pruning are not bidirectionally compatible, \textit{i.e.} video-specific methods struggle to adapt to the single-frame static image.
In comparison, the echo reconstruction mechanism in our temporal modeling is also applicable in image token selection. As shown in Tab.~\ref{tab: llava_img}, EchoPrune achieves competitive performance on three open-ended benchmarks: GQA \citep{hudson2019gqa}, SQA$_\mathrm{Img}$ \citep{lu2022learn} and TextVQA \citep{singh2019towards}, based on LLaVA-1.5-7B. The comparison methods include coverage-based MI-Pruner \citep{li2026mi}, attention-based VisionZip \citep{yang2025visionzip} and reconstruction-based ApET \citep{ma2026apet}. Specifically, MI-Pruner employs crossmodal and intra-modal mutual information as the scoring function, aligning with the Maximum Marginal Relevance framework as our approach. Similarly, ApET utilizes linear approximation in LLMs to filter out redundant tokens based on $L_2$ reconstruction error, aligning with our motivation.
While the literature on image pruning is extensive, we specifically focus on these representative methods due to their close methodological relevance. 
For a fair comparison with ApET, we initiate the reconstruction from $M{=}10$ tokens. The temperature $\tau{=}0.1$ is the same as LLaVA-OV in the main paper. 
As shown in the table, \ech~ achieves the highest average scores and robust performance on both video and image pruning tasks.

\begin{table}[h]
\centering
\caption{\textbf{Performance on LLaVA1.5-7B (Image-QA).} The vanilla model contains 576 tokens, which are compressed to 128 (22.2\% retained) and 64 (11.1\% retained).}
\label{tab: llava_img}
\resizebox{0.99\textwidth}{!}{
\begin{tabular}{l| cccc|cccc} 
\hline
\textbf{Method} & \textbf{GQA} & \textbf{SQA$_\mathrm{Img}$}  & \textbf{TextVQA} & \textbf{Average} & \textbf{GQA} & \textbf{SQA$_\mathrm{Img}$}  &\textbf{TextVQA}  & \textbf{Average} \\
\hline
\rowcolor{lightgray}
& \multicolumn{4}{c}{\textit{keep 128}} & \multicolumn{4}{c}{\textit{keep 64}} \\
\mi~\citep{li2026mi} \vv{2026.04} 
&  58.5 &  69.5 &  56.1  &  61.4 &  56.9  &  \sota{69.8}  & 54.9 & 60.5 \\
VisionZip \citep{yang2025visionzip}\vv{CVPR'25} 
&  57.6 &  68.9 &  56.8  &  61.1 &  55.1  &  69.0  &  55.5 & 59.9 \\
ApET \citep{ma2026apet} \vv{CVPR'26} 
&  58.9 &  68.7 &  53.9  &  60.5 &  56.9  &  68.9  &  53.0 & 59.6 \\
\rowcolor{lightpink}
\ech 
&  \sota{59.1} &  \sota{69.7} &  \sota{57.1}  &  \tbd{62.0} &  \sota{57.6}  &  68.5  &  \sota{55.6} & \tbd{60.6} \\
\hline
\end{tabular}
}
\end{table}

\begin{figure}[b]
\centering
\includegraphics[width=\linewidth]{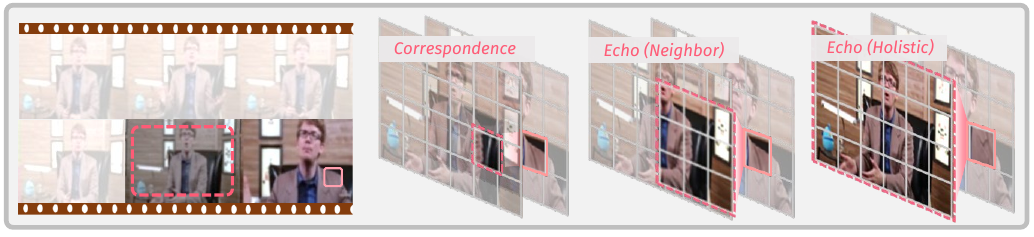}
\caption{
\textbf{Frame matching.}
\textit{Correspondence} (left) refers to the history patch in the same location. \textit{Neighbor} (middle) and \textit{Holistic} (right) of echo matching leverage the patches in a neighbor region or from the full frame, respectively.}
\label{fig:match}
\end{figure}

\subsection{Extended Visualization}
\label{subsec:app_more_vis}

Our reconstruction error is derived from the synergy of correspondence and echo matching.
As described in Sec.~\ref{subsec:exp_setup}, echo matching can be implemented in a neighbor region $\Omega_i^\epsilon$ (for LLaVA-OV), and in full-frame (for Qwen2.5VL- and Qwen3VL-series). We illustrate different ranges of frame matching in
Fig.~\ref{fig:match}. The choice of receptive field for echo matching is primarily dictated by the underlying vision encoder. LLaVA-OV adopts SigLip \citep{zhai2023sigmoid} and Qwen2 \citep{yang2024qwen2} as the vision encoder and LLM decoder. It needs more spatial constraints for its fixed resolution, \textit{i.e.} 196 tokens per frame. As a result, we conduct echo matching in a neighborhood for spatial consistency.
In comparison, Qwen2.5VL \citep{qwen2.5-VL} introduces dynamic resolution processing and absolute time encoding, enabling it to process images of varying sizes and videos of extended durations. To provide more contextual information, we employ holistic echo matching for QwenVL-series.

\begin{figure}[h]
\centering
\includegraphics[width=\linewidth]{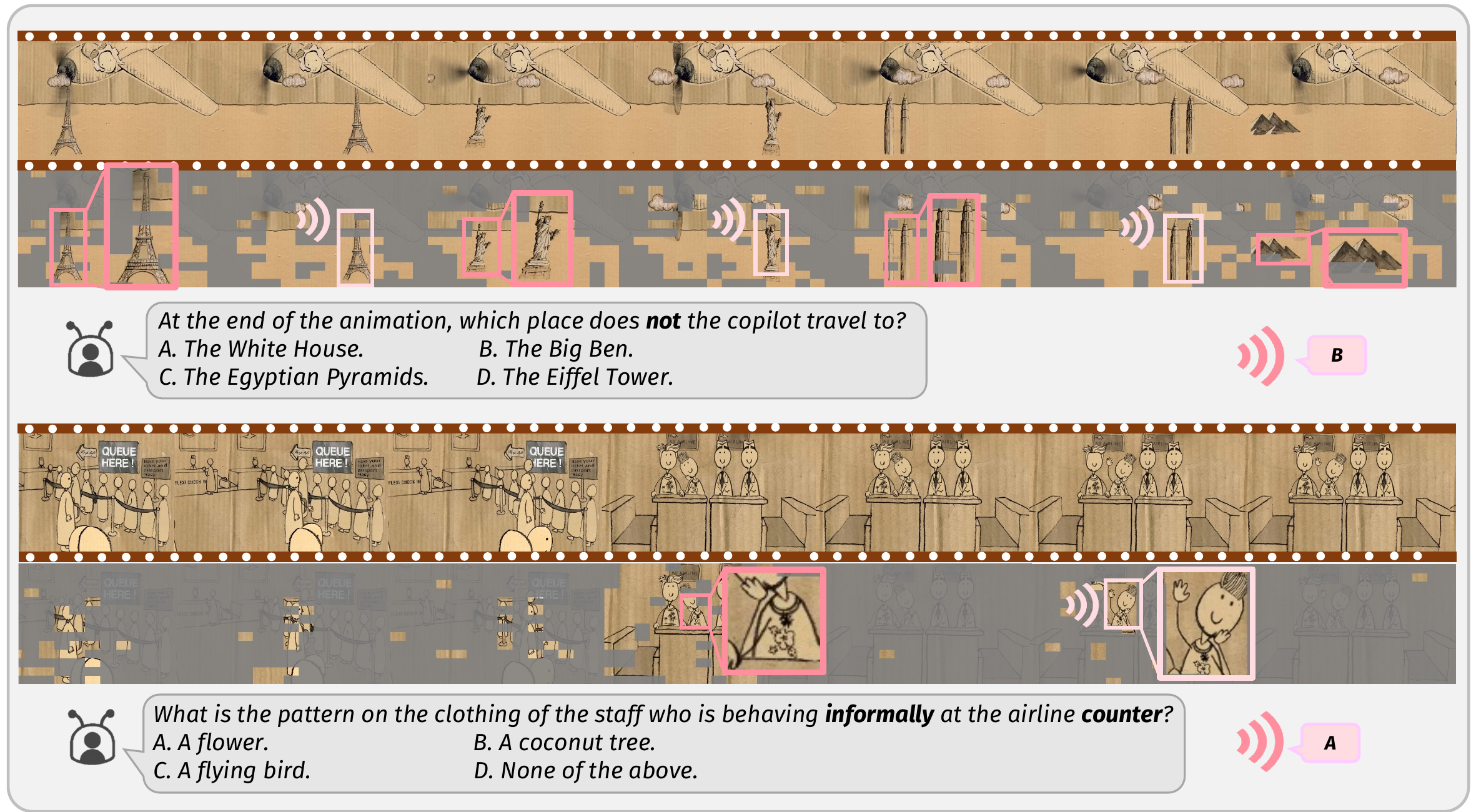} 
\caption{
\textbf{Pruning effects on LLaVA-OV-7B} (VideoMME, \textit{F160/20\%}). 
Our pruning effectively eliminates query-irrelevant patches while preserving pivot slots. Specifically, we track landmarks consistently \textbf{(top)} and identify new events of varying head directions \textbf{(bottom)}.
}
\label{fig:vis_llava_ii}
\end{figure}

We provide further visualizations of our adaptive pruning effects on LLaVA-OV in Fig.~\ref{fig:vis_llava_ii}, featuring two distinct questions based on the same video clip. When querying specific landmarks, EchoPrune effectively tracks the visual targets across multiple frames. The second question requires fine-grained retrieval regarding a staff member's clothing pattern, which appears in four frames. By discarding predictable tokens with minimal reconstruction error, EchoPrune successfully prunes the redundant frames. Specifically, among the four frames containing the staff (second row), only the first and third are retained, which correspond to distinct moments when the staff's head turns left and right, respectively. This shows that our method captures the new event with unpredictable information.

\end{document}